\pdfoutput=1
\documentclass[journal]{IEEEtran}
\usepackage[pdftex]{graphicx}
\usepackage{subfigure}
\usepackage{amsmath,amssymb,amsfonts}
\usepackage{color}
\usepackage{blindtext}
\usepackage{cite}
\usepackage{amsthm}
\usepackage[normalem]{ulem}
\usepackage{url}
\usepackage{comment}
\usepackage{multirow}
\usepackage{xcolor}
\usepackage{diagbox}
\usepackage{indentfirst}
\usepackage{array}
\usepackage[ruled]{algorithm2e}
\usepackage{algorithmic}
\usepackage{longtable}
\usepackage{hyperref}
\usepackage{xcolor}
\usepackage{graphicx}
\usepackage{booktabs}
\usepackage{calrsfs}
\usepackage{microtype}
\usepackage[english]{babel}
\usepackage[utf8x]{inputenc}
\usepackage{amsmath}
\usepackage{graphicx}
\usepackage[colorinlistoftodos]{todonotes}
\usepackage{textcomp}
\usepackage{gensymb}

\graphicspath{{./Figures/}}

\newcounter{gaocomm} 
\newcounter{Note}  
\definecolor{blue-violet}{rgb}{0.54, 0.17, 0.89}
\definecolor{mygreen}{rgb}{0.0, 0.5, 0.0}
\definecolor{awesome}{rgb}{1.0, 0.13, 0.32}
\definecolor{bostonuniversityred}{rgb}{0.8, 0.0, 0.0}

\newcommand{\GaoC}[1]{\textcolor{blue-violet}{\stepcounter{gaocomm}{\bf [Junbin's comment \arabic{gaocomm}: #1]}\;}}

\begin{document}

\title{\textbf{LSTM-Assisted Evolutionary Self-Expressive Subspace Clustering}}
\author{Di Xu, Tianhang Long and Junbin Gao
\thanks{ 
Di Xu and Junbin Gao are with the Discipline of Business Analytics, The University of Sydney Business School, The University of Sydney, NSW 2006, Australia. \protect E-mail: dixu3140@uni.sydney.edu.au; junbin.gao@sydney.edu.au
} 
\thanks{Tianhang Long is with Beijing Advanced Innovation Center for Future Internet Technology,
Beijing  Municipal  Key  Lab  of  Multimedia  and  Intelligent  Software  Technology,  Faculty  of  Information
Technology,  Beijing  University  of  Technology,  Beijing  100124,  China.  E-mail:  long54482000@163.com}
}

\maketitle

\markboth{IEEE Transactions on Neural Networks and Learning Systems, Vol.~XX, No.~X, XXX~2020
}%
{Xu, Long and Gao: LSTM For Evolutionary Self-Expressive Subspace Clustering}


\begin{abstract}
Massive volumes of high-dimensional data that evolves over time is continuously collected by contemporary information processing systems, which brings up the problem of organizing this data into clusters, i.e. achieve the purpose of dimensional deduction, and meanwhile learning its temporal evolution patterns. In this paper, a framework for evolutionary subspace clustering, referred to as LSTM-ESCM, is introduced, which aims at clustering a set of evolving high-dimensional data points that lie in a union of low-dimensional evolving subspaces. In order to obtain the parsimonious data representation at each time step, we propose to exploit the so-called self-expressive trait of the data at each time point. At the same time, LSTM networks are implemented to extract the inherited temporal patterns behind data in an overall time frame. An efficient algorithm has been proposed based on MATLAB. 
Next, experiments are carried out on real-world datasets to demonstrate the effectiveness of our proposed approach. And the results show that the suggested algorithm dramatically outperforms other known similar approaches in terms of both run time and accuracy.

\end{abstract}




\IEEEpeerreviewmaketitle
\section{Introduction}\label{Sec:1}
The recent decade has witnessed a gigantic explosion of data availability from various modalities and sources due to the advance of contemporary information processing systems. For instance, billions of cameras get installed worldwide and are ceaselessly generating data. This has promoted remarkable progresses and meanwhile created new challenges on how to acquire, compress, store, transmit, and process massive amounts of high-dimensional complex data. High dimensionalities of data will bring in severe computational as well as memorial burdens and mostly cut down the performance of existing algorithms. An important unsupervised learning problem encountered in such settings deals with finding informative parsimonious structures characterizing large-scale high-dimensional datasets. This task is critical for the detection of meaningful patterns in complex data and enabling accurate and efficient clustering. 

Based on the observation that though the collected dataset is in high dimension, the intrinsic dimension itself is commonly much smaller than that of its ambient space. In computer vision, for example, the number of pixels in an image could be rather large, yet most computer vision models use only a few parameters to describe the appearance, geometry, and dynamic of a scene. And this fact has motivated the development of lots of techniques for finding a low-dimensional representation of a high-dimensional dataset for the purpose of detecting underlying data characteristics. Traditional approaches, e.g. principle component analysis (PCA), are under the assumption that data is sampled from one single low-dimensional subspace of a high-dimensional space. In reality, the data points could be yet drawn from a number of subspaces and the membership of the data points to their corresponding subspaces is probably unknown. For example, a video sequence might have several moving objects, and different subspaces might be needed to describe the motions of their corresponding objects in the scene. Thus, we need to concurrently cluster the data into several subspaces as well as detect out a low-dimensional subspace fitting each group of points. This problem has found a wide range of applications in motion segmentation and face clustering in computer vision \cite{vidal2008multiframe}
, image representation and compression in image processing \cite{li2017robust}
, hybrid system identification in systems theory \cite{zheng2017consensus}
, robust principal component analysis (RPCA) \cite{wright2009robust}, and robust subspace recovery and tracking \cite{li2015learning}.

In the aforementioned settings, apart from the property that data could be reckoned as a bunch of points lying in a union of low-dimensional subspaces, data is also acquired at multiple points in time mostly. Thus, exploring its inherent temporal behaviors will obtain more information and higher clustering accuracy. For instance, we all know that feature point trajectories are related to motions in a video lying in an affine subspace \cite{tron2007benchmark}. Motions in any given short time interval are associated with motions in their latest past. Thus, aside from the union-of-subspaces structure in a video data, the underlying evolutionary structure is as well the key to characterize motions. As a result, designing proper frameworks that are capable of exploiting both the union-of-subspaces and temporal smoothness structures so as to conduct fast and precise clustering, especially in real-time cases where clustering is conducted in an online mode and a solution is required at each time step, is of great interest.

In recent years, researchers have been engaged in using deep networks to deal with various categories of temporal data and better performance has been reported in \cite{cao2018interactive, laptev2017time}. Among the existing neural network architectures, recurrent neural networks (RNNs), especially LSTM, have been proved to have even more superior performance than those of other networks and, not to mention, the commonly used traditional models. Meanwhile, conducting unsupervised clustering with LSTM structure has also been explored in diverse settings, See \cite{wollmer2009data} for details. 

Nonetheless, till now, few neural-networks (NN) related algorithms have covered the topic of evolutionary subspace clustering. In the meantime, most researchers focus on exploring and improving the schemes of static subspace clustering \cite{vidal2011subspace} with solely a few studies taking advantages of the intrinsic temporal patterns within data. Recently, Hashemi and Vikalo \cite{hashemi2018evolutionary} investigated both the evolutionary and a-union-of-subspaces properties of a motion segmentation problem. They introduced a convex evolutionary self-expressive model (CESM), an optimization framework that exploits the self-expressiveness property of data and learns sparse representations while taking into account prior representations. However, their processing of temporal patterns in data only limits in implementing classic weighted average smoothing techniques, which lead to sub-optimal efficiency in extracting complex information of data evolution. 

In this paper, we further explore the idea of CESM and extend their design by utilizing LSTM networks to learn the evolutionary self-expressive data representation. Our framework alternates between two steps - LSTM clustering and spectral clustering - to study both the data evolution and segmentation. The advantages of the proposed scheme is in two folds; not only we implement deep networks in seeking the temporal patterns as well as subspace representation of data, but we obtain as well a notably faster and more accurate algorithm compared to the past models.

The rest of this paper is organized as follows. Section \ref{Sec:2} overviews existing approaches to subspace clustering, evolutionary clustering, and evolutionary subspace clustering. Section \ref{Sec:3} proposes our LSTM evolutionary self-expressive subspace clustering framework. Section \ref{Sec:4} describes the implementation of this scheme in MATLAB programming. Section \ref{Sec:5} presents experimental results on real-world datasets. Then, the concluding remarks are stated in section \ref{Sec:6}.

\section{Background}\label{Sec:2}
 In this section, we first state the notation used all through the paper. Next, we recap multiple past subspace clustering, evolutionary clustering and evolutionary subspace clustering methods. At the end, we spotlight distinctive features of the evolutionary subspace clustering framework that we will introduce in the succeeding sections.

\subsection{Notation}\label{Sec:2_1}

Bold capital letters denote matrices such as $\mathbf X$ while bold lowercase letters such as $\boldsymbol{x}$ represent vectors. Sets as well as subspaces are denoted by calligraphic letters, such as $\mathcal{X}$ for a set and $\mathcal{S}$ for a subspace. 
$N$ is the number of data points in a set, e.g. $\mathcal{X}$, while $D$ is the corresponding actual dimension of that set. $n$ is the total number of subspaces that the points of a set $\mathcal{X}$ belong to, with $\{\mathcal{S}_i\}_{i=1}^n$. And the dimension of a subspace $\mathcal{S}$ is denoted by $d$. $T$ is the total number of time steps in an evolving process while $t$ is a specific timestep constrained by $1 \leq t \leq T$.  Let $\{\mathbf{X}_t\}_{t=1}^T$ be a time series in which each $\mathbf X_t$ collects all the features of objects at time $t$. They may have different structures, such as matrices or tensors, depending on the context of specific application scenarios.  Denote by $\mathbf{I}$ is the identity matrix of appropriate sizes. Further, $\|\mathbf{X}\|_*$ denotes the nuclear norm of $\mathbf{X}$ defined as the sum of singular values of $\mathbf{X}$. $\text{diag}(\mathbf{X})$ outputs the diagonal matrix of the diagonal elements in $\mathbf{X}$. $\text{tr}(\mathbf{X})$ yields the trace of the matrix $\mathbf{X}$.  $\text{vec}(\mathbf{X})$ constructs a vector by orderly stitching all the columns of $\mathbf{X}$ together. Finally, $\text{sign}(x)$ returns the sign of its argument and $\text{ceil}(x)$ returns the integer rounding up to its argument.

\subsection{Subspace Clustering} \label{Sec:2_2}
Subspace clustering has been notably highlighted over the past two decades (see, e.g., \cite{RVidal2011} and the references therein). The motivation behind is to arrange data into clusters so as to find a union of subspaces that the data points are drawn from. In specific, let $\{\boldsymbol{x}_j \in \mathbb{R}^D\}_{j=1}^N$ be a given set of points drawn from a unknown union of ${n \ge 1}$ linear subspace 
$\{\mathcal{S}_i\}_{i=1}^n$ 
of unknown dimensions $d_i=\text{dim}(\mathcal{S}_i)$, ${0<d_i<D}$, $i=1, ..., n$. Suppose that $\mathbf{U}_i$ is a subspace basis which is to be identified in most learning tasks.  

When the number of subspaces is equal to $1$, this problem of searching for $\mathbf U_i$ from a set of noised data $\{\mathbf x_j\}$ comes into the widely-applied PCA \cite{jolliffe2003principal}. The ultimate goal of subspace clustering is to find parameters $n, \{d_i\}_{i=1}^n$ and $\{\boldsymbol{U}_i\}_{i=1}^n$ 
described above as well as the segmentation of data points with respect to subspaces \cite{RVidal2011}. In the case of affine subspaces learning, except for the above parameters to be learned, we also need to identify a translation vector $\boldsymbol{\mu}_i$ for each affine subspace. This paper is concerned with the linear subspace clustering for which we assume that all $\boldsymbol{\mu}_i= \mathbf 0$.

Existing studies of subspace clustering can be divided into four main categories: (i) Algebraic;  (ii) Iterative; (iii) Statistical and (iv) Spectral Clustering methods. Algebraic methods, such as \cite{vidal2008multiframe}, are mostly designed for linear subspaces and not robust to noise. 
Iterative refinement is a relatively efficient technique in improving the performance of algebraic methods on noisy data, whereas, its prominent K-planes algorithm \cite{bradley2000k} suffers from a rigid requirement on initialization and severe sensitivity to outliers. 
Statistical approaches \cite{derksen2007segmentation} further improve the sub-optimal results in the aforesaid two categories by specifying an appropriate generative model for data. Nevertheless, this class of methods on the whole have theoretical imperfection and further improvement is needed. 


Spectral clustering \cite{ng2002spectral} 
is a highly desirable technique for clustering high-dimensional datasets. 
However, one of the main challenges in applying spectral clustering to subspace clustering problems is to define a decent affinity matrix. Since two data points with a close physical distance do not guarantee to lie in the same subspace (take the points near intersections as an instance) and vice versa. As a consequence, typical distance-based affinity measurements are no longer appropriate in this case.

To obtain better clustering outcomes for subspace clustering, researchers have done plenty of trials in constructing an appropriate pairwise affinity matrix for points lying in multiple subspaces, such as the factorization and GPCA -based affinity \cite{vidal2008multiframe, tron2007benchmark}, local subspace affinity (LSA) \cite{yan2006general} and spectral local best-fit flats (SLBF) and locally linear manifold clustering (LLMC) \cite{goh2007segmenting}. However, they still do not well handle the data points near intersections and will not work properly when it comes to dependent subspaces.

Lately, approaches based on spectral clustering which assume data is self-expressive \cite{elhamifar2009sparse} and drawn from a union of subspaces to form the affinity matrix, which is obtained by solving an optimization problem that incorporates $\ell_1$, $\ell_2$ or nuclear norm regularization, have become exceedingly trendy. The self-expressiveness property states that each data point can be represented by a linear combination of other points in the same   subspaces at the data point. Mathematically, this can be represented as
\begin{equation}
\mathbf{X} = \mathbf{X}\mathbf{C} \;
,\;
\text{diag}(\mathbf{C})=0 \
\label{Eq:Aa1}
\end{equation}
where $\mathbf{X}=[\boldsymbol{x}_1, ..., \boldsymbol{x}_N]\in\mathbb{R}^{D\times N}$ is the data matrix and $\mathbf{C}=[\boldsymbol{c}_1, ..., \boldsymbol{c}_N]\in\mathbb{R}^{N\times N}$ is the matrix of coefficients, sometimes called the affinity matrix of the dataset. We will use the name either coefficient matrix or affinity matrix.

Ideally, such an affinity matrix shall demonstrate the so-called ``block'' structure if the dataset has been arranged orderly according to its belonging subspaces. Based on this ideal fact, in the core, all the self-expressive subspace clustering approaches manage to handle variants of the optimization problem, which is
\begin{equation}
\min_{\mathbf{C}} \|\mathbf{C}\|
\; \;
s.t.
\; \;
\|\mathbf{X}-\mathbf{X}\mathbf{C}\|\leq \xi
,\; \;
\text{diag}(\mathbf{C})=0.
\label{Eq:Aa2}
\end{equation}
Given $\mathbf{C}$, one can build a matrix $\mathbf{A}$ as $\mathbf{A}=|\mathbf{C}|+|\mathbf{C}|^T$, and then apply spectral clustering on $\mathbf{A}$ to cluster data.  

In order to establish an affinity matrix $\mathbf{C}$, the sparse subspace clustering (SSC) algorithm \cite{elhamifar2009sparse} proceeds by implementing the basis pursuit (BP) algorithm. 
Least squares regression (LSR) \cite{lu2013correlation} employs $l_2$ regularization to find $\mathbf{C}$. Nuclear norm minimization is applied to low rank subspace clustering (LRSC) \cite{liu2013robust} to get $\mathbf{C}$. Feng \emph{et al.} \cite{feng2014robust} pursues the block-diagonal structure of $\mathbf{C}$ by deriving a graph Laplacian constraint based formulation and then proposes a stochastic sub-gradient algorithm for optimization. Gao \emph{et al.} \cite{gao2015multi} proposes multi-view subspace clustering which conducts clustering on the subspace representation of every view simultaneously. In \cite{nie2016subspace}, low-rank representation learning and data segmentation are jointly processed by searching for individual low-rank segmentation as well as implementing the Schatten $p$-norm relaxation of the non-convex rank objective function. Lastly, \cite{7222444} obtains the similarity matrix by thresholding the correlations between data points. Lately, further generalizations of SSC and Low rank representation (LRR) schemes are presented. In specific, \cite{li2017structured} suggests a SSC-based method that jointly conducts clustering and representation learning. The SSC was generalized to handle data with missing information \cite{elhamifar2016high,tsakiris2018theoretical}.

Li \emph{et al.}  \cite{li2015temporal} propose the temporal subspace clustering algorithm which samples a single data point at each time step and aims at assembling data points into sequential segments, and then come after by clustering the segments into their corresponding subspaces. 



\subsection{Evolutionary Clustering}\label{Sec:2_3}
Aforementioned works are under the assumption that data is obtained in an offline mode and will be fixed without any evolution through time once acquired.

The discussion of evolutionary clustering has been trending in the past few years and related techniques have been widely applied in a range of practical settings \cite{Chakrabarti:2006:EC:1150402.1150467, Chi:2007:ESC:1281192.1281212}.
Under the assumption that immediate changes of clustering in a short period of time are not desirable, evolutionary clustering is the problem of organizing timestamped data to generate a clustering sequence by introducing a temporal smoothness framework. 
A high-quality evolutionary clustering algorithm should be capable of well fitting the data points at each time step as well as generating a smooth cluster evolution that can provide the data analyst with a coherent and easily explainable model. 

Chakrabarti \emph{et al.} \cite{Chakrabarti:2006:EC:1150402.1150467} originally propose a generic framework for evolutionary clustering by adding a temporal smoothness penalty term to a static clustering objective function, and this general framework explores K-means and agglomerative hierarchical clustering as illustrative examples. 
Chi \emph{et al.} \cite{chi2007evolutionary,chi2009evolutionary} further expand on this idea by suggesting an evolutionary spectral clustering approach. It constructs the following loss function (\ref{Eq:Ab1}) that contains both consistency and smoothness terms
\begin{equation}
\mathbf{L} = \alpha \mathbf{L}_\text{temporal} + (1-\alpha)\mathbf{L}_\text{snapshot} 
\label{Eq:Ab1}
\end{equation}
where $\mathbf{L}_\text{snapshot}$ symbolizes the term of consistency, i.e. the current spectral clustering loss, $\mathbf{L}_\text{temporal}$ symbolizes that of smoothness, and $\alpha$ is a smoothing parameter to decide the weight of current spectral clustering loss. Equation (\ref{Eq:Ab1}) presumes a certain degree of temporal smoothness between $\mathbf{X}_{t-1}$ and $\mathbf{X}_{t}$, where the smoothness has the capability of preserving either the Cluster Quality (PCQ) or Cluster Membership (PCM) \cite{chi2007evolutionary,chi2009evolutionary}.

Rosswog \emph{et al.} \cite{rosswog2008detecting} proposes evolutionary extensions of K-means as well as agglomerative hierarchical clustering by filtering the feature vectors using a finite impulse response (FIR) filter which aggregates the estimations of feature vectors. The affinity matrix is then computed among the filtered outcomes instead of the feature vectors. 
Also based on the idea of adjusting similarities followed by static clustering, AFFECT algorithm, as an extension to static clustering, is proposed in \cite{xu2014adaptive}, where the similarity matrix at a specific time $t$ is assumed as the sum of a deterministic matrix, i.e. the affinity matrix, and a Gaussian noise matrix. Nonetheless, for the sake of searching for an optimal smoothing factor $\alpha_t$, AFFECT makes rather strong assumptions on the structure of affinity matrices , which is generated by assuming a block structure for affinity matrix that only stands when the data at each time step $t$ is a realization of a dynamic Gaussian mixture model. But this is generally not true in practice.

\subsection{Evolutionary Subspace Clustering}\label{Sec:2_4}
For the sake of conducting evolutionary clustering at multiple time steps to temporal data identified by a union-of-subspaces structure, researchers probably will think about chaining snapshots from all the time steps together and applying one specific subspace clustering technique w.r.t. this set \cite{hashemi2018evolutionary}. Nevertheless, the concatenation will induce a dramatic rise in feature numbers and, thus, lead to unwanted increases in computational complexity. Moreover, by fitting the set with a single union of subspaces, it's almost infeasible to uncover the slight evolutionary changes in the temporal structure of data, which results in an inefficient temporal evolution of subspaces.

We will consider the real-time motion segmentation task \cite{smith1995asset} as an illustration over here, where the aim of this task is to recognize and track motions in a video sequence. The problem of real-time motion segmentation is associated with that of offline motion segmentation \cite{elhamifar2009sparse}. The difference between them is that by taking use of all the frames in the sequence, clustering is performed only once in the offline scenario, whereas by receiving each snapshot of the sequence one at a time, clustering is performed step by step in the online setting. The generated subspaces depict the evolution of motions, in which subspaces in closer snapshots are similar while in more far-apart ones may significantly deviate. Thus, as in the aforesaid C\&C approach, imposing a sole subspace layout for the whole time sequence may incur poor clustering outcomes. And a regime that is capable of judiciously taking advantages of the evolutionary structure while preserving the union-of-subspaces structure is in need.

Instead of assuming a common subspace for all clusters, Vahdat \emph{et al.} \cite{vahdat2010bottom,vahdat2012symbiotic,vahdat2014evolutionary} proposes a bottom-up symbiotic evolutionary subspace clustering (S-ESC) algorithm. 
In \cite{xu2014adaptive,Chi:2007:ESC:1281192.1281212,chi2009evolutionary,rosswog2008detecting}, authors propose evolutionary subspace clustering techniques by employing static clustering algorithms, such as spectral clustering, etc., to process the affinity matrix and then apply equation (\ref{Eq:Ac1}) for evolutionary smoothing. 
\begin{equation}
    \mathbf{C}_t = {\alpha}_t\bar{\mathbf{C}}_t + (1-{\alpha}_t)\mathbf{C}_{t-1}
    \label{Eq:Ac1}
\end{equation}
where 
$\bar{\mathbf{C}}_t$ and $\mathbf{C}_{t-1}$ denote the affinity matrix constructed solely from $\mathbf{X}_t$ and smoothed affinity matrix at time $t-1$ respectively, and ${\alpha}_t$ is the smoothing parameter at time $t$. It is worthwhile mentioning that AFFECT algorithm \cite{xu2014adaptive} provides a procedure for finding the smoothing parameter ${\alpha}_t$.


Lately, Hashemi \emph{et al.} \cite{hashemi2018evolutionary} propose an ESCM framework that exploits the self-expressiveness property of data to learn a representation for $\mathbf{X}_t$ and meanwhile takes into account of data representation learnt in the previous time step by the following procedure, 
\begin{equation}
    \mathbf{C}_t = f_\theta (\mathbf{C}_{t-1}),\;\;\; \mathbf{X}_t = \mathbf{X}_t\mathbf{C}_t,\;\;\; \text{diag}(\mathbf{C}_t) = \mathbf{0},
\label{Eq:Ac2}
\end{equation}
where
$    f_\theta (\mathbf{C}_{t-1}) \in \mathcal{P}_C$ is a parametric family of clustering representations. 
In theory, the function $f_\theta : \mathcal{P}_C \rightarrow \mathcal{P}_C$ 
could be any parametric function while the set $\mathcal{P}_C \subseteq \mathbb{R}^{ N_t \times N_t}$ stands for any preferred parsimonious structure imposed on the representation matrices at each time instant, e.g., sparse or low-rank representations. 

Hashemi \emph{et al.} \cite{hashemi2018evolutionary} recommended  
\begin{equation}
    \mathbf{C}_t = f_\theta (\mathbf{C}_{t-1}) = \alpha \mathbf{U} + (1-\alpha)\mathbf{C}_{t-1},
    \label{Eq:Ac3}
\end{equation}
where the values of parameters $\theta = (\mathbf{U},\alpha)$ to be learned specify the relationship between $\mathbf{C}_{t-1}$ and $\mathbf{C}_t$, and the innovation representation matrix $\mathbf{U}$ captures changes in the representation of data points between consecutive time steps. 

Though \cite{hashemi2018evolutionary} has done a successful trial of evolutionary subspace clustering with its proposed ESCM framework, merely employing (\ref{Eq:Ac3}) on time-series smoothing makes the whole model lack of expressiveness and unable to efficiently unveil the substantial evolutionary information behind data. In this paper, to address the aforementioned issues, we propose to substitute (\ref{Eq:Ac3}) with recurrent LSTM deep networks for the reason of significant performance of LSTM in learning complicate temporal patterns under various settings. As the experimental results demonstrate, our proposed framework does effectively capture the temporal behaviors of data and achieve a significantly better accuracy than previous ones, which further prove the fitting and capability of LSTM networks in processing temporal information.

\section{LSTM Evolutionary Subspace Clustering}\label{Sec:3}
In this section, we will present our model for the evolutionary subspace clustering based on the recurrent neural networks LSTM.

\subsection{The Proposed Framework based on LSTM}
Let $\{\mathbf{x}_{t,i}\}_{i=1}^{N_t}$ be a set of real-valued $D_t$ dimensional vectors at time $t$, and we aggregate them into a matrix as $\mathbf{X}_t = [\mathbf{x}_{t,1}, ..., \mathbf{x}_{t,N_t}] \in \mathbb{R}^{D_t \times N_t}$.  
All the data points at time step $t$ are drawn from a union of $n_t$ evolving subspaces $\{\mathcal{S}_{t,i}\}_{i=1}^{n_t}$ with dimensions $\{d_{t,i}\}_{i=1}^{n_t}$. And the full dataset is $\mathcal{D} = \{\mathbf X_t\}^T_{t=1}$. Without a loss of generality, we assume that the columns of $\mathbf{X}_t$ are normalized vectors with unit $l_2$ norm. Due to the underlying union of subspaces structure, the data points at each time step themselves satisfy the self-expressiveness property \cite{elhamifar2009sparse} formally stated in equation (\ref{Eq:Aa1}) of Section \ref{Sec:2_2}.

The main purpose of most existing static subspace clustering algorithms is to partition $\{\mathbf{x}_{t,i}\}_{i=1}^{N_t}$ into $n_t$ groups so as to allocate data points that belong to the same subspace into the same cluster. Manipulating the self-expressive characteristic of data does improve the progress of exploiting the fact that a collection of data points belongs to a union of subspaces. 

In numerous applications of subspace clustering, apart from lying in a union of subspaces, data is also of temporal patterns, which makes researchers get started on exploring the self-expressiveness based evolutionary subspace clustering. Still, the past studies focus on applying weighted average akin methods or other relatively simple temporal processing models to smooth the temporal clustering results at multiple time steps \cite{hashemi2018evolutionary}, with less people taking advantage of the capability of deep learning methods in learning data evolution.  

Upon their chain-like nature, recurrent neural networks (RNNs) have been showed to have excellent performance on persisting past information, which naturally works for scenarios of processing various types of sequences. Further, Long short term memory (LSTM), a very special kind of RNNs well capable of learning long term dependencies, works much better than the standard version of RNNs for many tasks\cite{sainath2015convolutional}. To this end, we propose to find a representation matrix $\mathbf{C}_t$ for each time $t$ using LSTM networks, such that $f_\theta$ in equation (\ref{Eq:Ac2}) is instead implemented by LSTM networks. We will refer to our proposed evolutionary subspace clustering scheme that satisfies equation (\ref{Eq:Ac2}), with $f_\theta$ expressed by LSTM networks, as LSTM evolutionary subspace clustering method (LSTM-ESCM) in what follows.

\begin{figure}[ht]
\includegraphics[width=\linewidth]{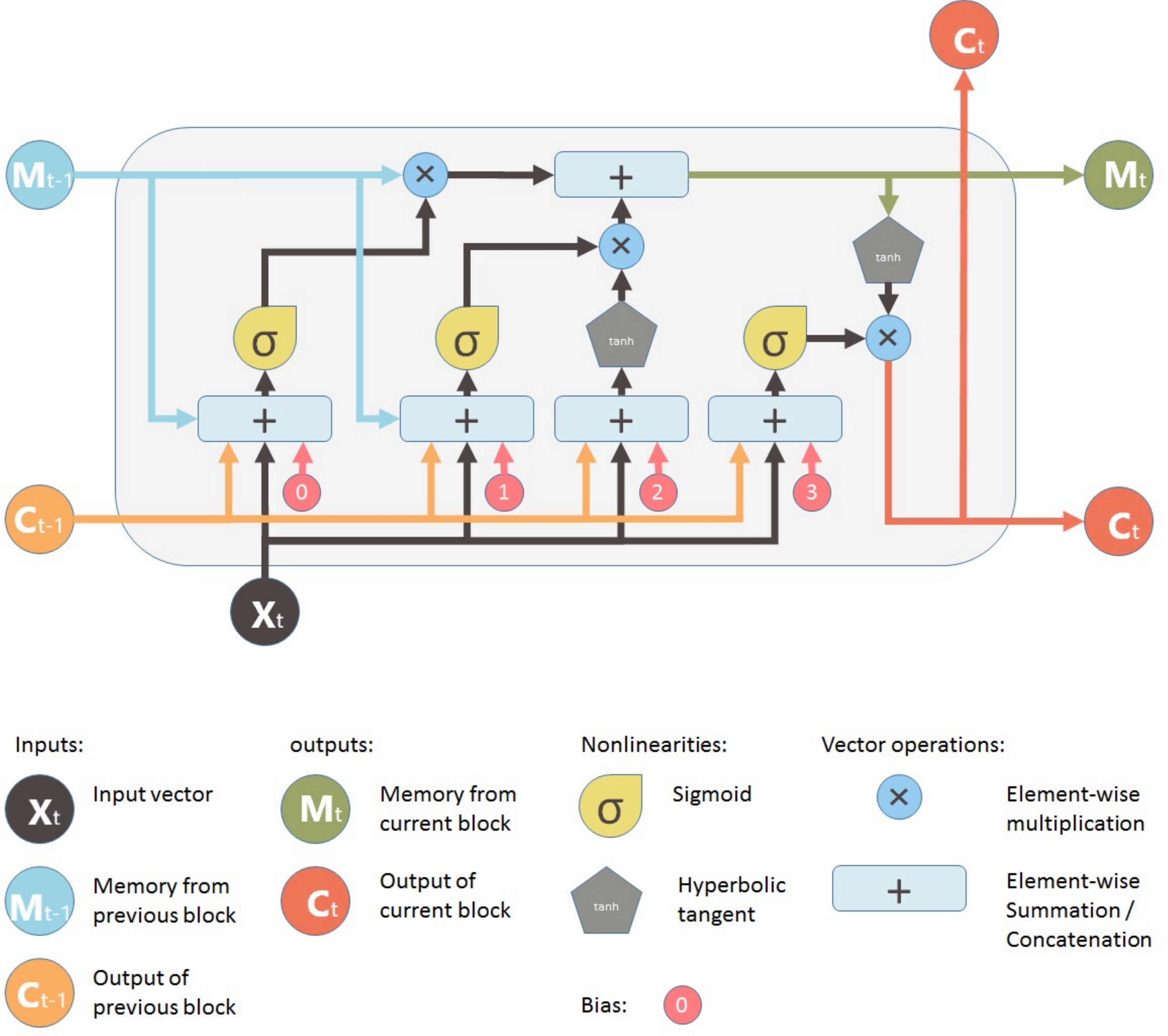}
\caption{LSTM network structure of LSTM-ESCM}
\label{Fig:B1}
\end{figure}

An LSTM block consists of three inputs - cell and hidden states generated from time step $t-1$ and data information at $t$, two outputs - processed cell and hidden states for timestamp $t$, and in between are multiple gates to optionally let information through. In our scenario, as shown in \textbf{Fig. \ref{Fig:B1}}, we use $\mathbf{M}_{t-1}$ and $\mathbf{M}_t$ to denote the cell states at time step $t-1$ as well as $t$, which represent the memories of previous and current blocks. $\mathbf{C}_{t-1}$ and $\mathbf{C}_t$ are hidden states, i.e. our smoothed clustering results, at time step $t-1$ and $t$. $\mathbf{X}_t$ is our input data at time point $t$. And then the proposed LSTM-ESCM is proceeded by equations in (\ref{Eq:B2}), where $\widetilde{\mathbf{M}}_t$, $f_t$, $i_t$, and $o_t$ are new cell state candidate values, forget, input, and output gates of current block at timestamp $t$ correspondingly.

\begin{equation}
\begin{aligned}
    & f_t = \sigma (W_f \cdot [\mathbf{C}_{t-1},\mathbf{X}_t] + b_f) \\
    &  i_t = \sigma (W_i \cdot [\mathbf{C}_{t-1},\mathbf{X}_t] + b_i) \\
    & \widetilde{\mathbf{M}}_t = \tanh (W_m \cdot [\mathbf{C}_{t-1}, \mathbf{X}_t] + b_m) \\
    & \mathbf{M}_t = f_t * \mathbf{M}_{t-1} + i_t * \widetilde{\mathbf{M}}_t \\
    & o_t = \sigma (W_o \cdot [\mathbf{C}_{t-1}, \mathbf{X}_t] + b_o) \\
    & \mathbf{C}_t = o_t * \tanh{(\mathbf{M}_t)}
\label{Eq:B2}
\end{aligned}
\end{equation}

According to the above LSTM, the hidden state $\mathbf{C}_t$ can be represented as a function of $\mathbf{C}_t = f_{\theta}(\mathbf{C}_{t-1}, \mathbf{X}_t)$ where $\theta$ denotes all the networks parameters $W$'ss and $b$'s. We propose to use the hidden state as the self-expressive matrix, hence the natural requirement according to equation \eqref{Eq:Aa1} prompts us to construct the following optimization regime
\begin{equation}
\begin{aligned}
    & \min_\theta  \frac12 \|\mathbf{X}_t - \mathbf{X}_t f_\theta (\mathbf{C}_{t-1},\mathbf{X}_t)\|_F^2, \\
    &\text{s.t.}  \;\;  
    \text{diag}(\mathbf{C}_t) = \mathbf{0}.
\label{Eq:B3}
\end{aligned}
\end{equation}
After getting a solution to (\ref{Eq:B3}) by solving the general constrained representation learning problem, we build an affinity matrix $\mathbf{A}_t = |\mathbf{C}_t| + |\mathbf{C}_t|^T$ and then apply spectral clustering to $\mathbf{A}_t$.

\subsection{Implementation of LSTM-ESCM}\label{Sec:4}

To implement the framework introduced in the last subsection, we shall carefully define the learning objective --- the loss function for the model.  The basic building block \eqref{Eq:B3} consider the consistent requirement at the timestamp $t$. Given that the hidden state variable $\mathbf{C}_t$ is being used as the data self-expressive matrix, we shall propose the following different models

\subsubsection{Model 1:}  In this model, we will train the proposed LSTM with the following loss function over the data sequence
\begin{align}
\min_{\theta} \mathbf{L} = \frac1{2T}\sum^T_{t=1}\|\mathbf{X}_t - \mathbf{X}_t \mathbf{C}_t\|^2_F, \;\;\; \text{s.t.}\;\; \text{diag}(\mathbf{C}_t) = \mathbf 0
\end{align}

We call this the Plain LSTM-ESCM.

\subsubsection{Model 2:} In the sparse subspace clustering, it is desired that the self-expressive matrix should be sparse. There are different ways to achieve this. For example, based on the deep learning approach, we can add an drop-out layer on the topic of hidden state output. However to have an intuitive implementation, we propose adding a sparse-inducing regularization to the loss function. This introduces out the so-called Sparse LSTM-ESCM
\begin{align}
\min_{\theta} \mathbf{L} = & \frac1{2T}\sum^T_{t=1}\|\mathbf{X}_t - \mathbf{X}_t \mathbf{C}_t\|^2_F +\lambda \sum^T_{t=1}\|\mathbf{C}_t\|_1, \notag\\
& \text{s.t.}\;\; \text{diag}(\mathbf{C}_t) = \mathbf 0.
\end{align}
where $\|\cdot\|_1$ is the $\ell_1$ norm of the matrix, equal to the sum of the absolute values of all the matrix elements. When $\lambda = 0$, this model goes back to the Plain LSTM-ESCM.  So we will only analyze the model Sparse LSTM-ESCM, i.e., LSTM-ESCM.



To conduct the backpropagation for training the LSTM with the new loss function, we shall work out the derivative of the loss function with respect to the hidden state variable $\mathbf{C}_t$.  Obviously, as shown in Figure~\ref{Fig:B1}, the output of our last layer is the self-expressive matrix $\mathbf{C}_t \in \mathbb{R}^{N_t \times N_t}$ while the target is $\mathbf{X}_t \in \mathbb{R}^{D_t \times N_t}$\footnote{As these dimensions of those two: input and output variable,  do not align, it is impossible to define a customized loss function in MATLAB deep learning toolbox.  But fortunately we have managed to rewrite the loss as the form that is compatible with the requirement.}. Specifically consider the loss term at timestamp $t$, 
\begin{align}
L &= \frac12 \|\mathbf{X}_t - \mathbf{X} \mathbf{C}_t\|^2_F +\lambda \|\mathbf{C}_t\|_1  
\notag\\
&= \frac12\text{tr}((\mathbf{I}-\mathbf{C}_t)^T \mathbf{X}_t^T \mathbf{X}_t (\mathbf{I}-\mathbf{C}_t)) +\lambda \|\mathbf{C}_t\|_1\notag\\
& = \frac12 \text{tr}((\mathbf{X}_t^T\mathbf{X}_t)((\mathbf{I}-\mathbf{C}_t)(\mathbf{I}-\mathbf{C}_t^T)))+\lambda \|\mathbf{C}_t\|_1\label{Eq:CA1}
\end{align}
and its corresponding derivative is
\[
\frac{\partial L}{\partial \mathbf C}_t =  (\mathbf{X}_t^T\mathbf{X}_t)(\mathbf{C}_t - \mathbf{I}) + \lambda\text{sign}(\mathbf C_t).
\]

%

We know the size of $\mathbf{X}_t^T\mathbf{X}_t \in \mathbb{R}^{N_t \times N_t}$, as new targets, is same as the size of $\mathbf{C}_t \in \mathbb{R}^{N_t \times N_t}$, as the output from the network.

\subsection{Data Structure in MATLAB Implementation}
We have re-written the loss function in such a way that the new target data $\mathbf{X}_t^T\mathbf{X}_t \in \mathbb{R}^{N_t \times N_t}$ has the same size as that  of $\mathbf{C}_t \in \mathbb{R}^{N_t \times N_t}$, as the output from the network. This has satisfied the requirement in MALTAB deep learning toolbox.

Another issue with MATLAB deep learning toolbox is that the LSTM layer only accepts vector sequence, thus we have to convert our inputs and targets into vectors.

At any time point $t$, we have a data matrix $\mathbf{X}_t$ containing features for $N_t$ objects (to be clustered) in columns while the rows corresponding to $D_t$ features of objects. That is $\mathbf{X}_t \in\mathbb{R}^{D_t\times N_t}$.  In the most experiments of this paper, we have assumed we always have $N_t$ objectis to be clustered at any time step. 

However if the number of objects at different timestamps is different, we suggest a full size LSTM should be constructed with the largest $N_0$.  If an $\mathbf X_t$ has less columns than $N_0$, we will randomly add $N_0-N_t$ columns of zero vectors. In calculating the loss and derivatives, we will take out those terms $c_{ij}$ where either $i$ or $j$ is one of column indices where zeros vectors are added. Intuitively we are not care about those similarity $c_{ij}$'s. More formally, we can see it in this way. For the sake of convenience, we assume that after adding zero columns we have $\widehat{\mathbf{X}}_t = [\mathbf{X}_t, \mathbf{0}]$. Denote $\mathbf{I} - \mathbf{C}_t = \begin{bmatrix}\mathbf{I}_{11} - \mathbf{C}_{11} & \mathbf{I}_{12} - \mathbf{C}_{12}\\\mathbf{I}_{21} - \mathbf{C}_{21} & \mathbf{I}_{22} - \mathbf{C}_{22}\end{bmatrix}$. Note that $\mathbf{I}_{12} = \mathbf{I}_{21}^T = \mathbf{0}$. Then for the loss over the LSTM, the term concerning us is
\begin{align*}
&\|\widehat{\mathbf{X}}_t - \widehat{\mathbf{X}}_t\mathbf{C}_t\|^2_F=\|\widehat{\mathbf{X}}_t (\mathbf{I} -\mathbf{C}_t)\|^2_F\\
=&\|\begin{bmatrix} {\mathbf{X}}_t(\mathbf{I}_{11} -\mathbf{C}_{11}) &  - {\mathbf{X}}_t \mathbf{C}_{12}\end{bmatrix}\|^2_F\\
=& \| {\mathbf{X}}_t(\mathbf{I}_{11} -\mathbf{C}_{11})\|^2_F + \| {\mathbf{X}}_t \mathbf{C}_{12}\|^2_F
\end{align*}
The first term is actually the loss for the data ${\mathbf{X}}_t$ without adding any zero columns, so we can remove the second term when calculating loss. Or except for $\mathbf{C}_{11}$, on the top of the hidden state, we can add a layer zeroing out all $\mathbf{C}_{12}, \mathbf{C}_{21}$ and $\mathbf{C}_{22}$.


For a full dataset $\mathcal{D}$, it is a general practice to break down the time series into shorter sections of sequence. For example, using a moving window to slide out a length $S_t$ shorter sequence, or cut the entire sequence into many shorter sequences without overlapping.  The number of such shorter sequences is denoted by $N$, i.e., the training number (or the total number of training and testing), and these sequences will be sent into networks  for training purpose individually. Our data is processed by the former option.

Under MATLAB syntax, we organize each ``data'' (a shorter sequence) in the following matrix
\[
\mathbf{X}^i = [\text{vec}(\mathbf{X}_{t_i+1}), ..., \text{vec}(\mathbf{X}_{t_i+s})] \in \mathbb{R}^{(D_t\times N_t) \times S_t}
\]
where $i = 1, 2, ..., N$. All these training data in matrix should be organized into MATLAB cells of size $N\times 1$, that is,
\[
\mathcal{X}_{\text{train}} = \{\mathbf{X}^1, ..., \mathbf{X}^N\}
\]

Given the special loss function defined in \eqref{Eq:CA1}, we organize the target data as
\[
\mathbf{Y}^i = [\text{vec}(\mathbf{X}^T_{t_i+1}\mathbf{X}_{t_i+1}), ..., \text{vec}(\mathbf{X}^T_{t_i+s} \mathbf{X}_{t_i+s})] \in \mathbb{R}^{N_t^2 \times S_t}
\]
and the target data $\mathcal{Y}_{\text{train}}$  will be organized into MATLAB cells of size $N\times 1$ as well, which is
\[
\mathcal{Y}_{\text{train}} = \{\mathbf{Y}^1, ..., \mathbf{Y}^N\}
\]

Similarly, the output clustering and affinity results for each sequence will be
\[
\mathbf{C}^i = [\text{vec}(\mathbf{C}_{t_i+1}), ..., \text{vec}(\mathbf{C}_{t_i+s})] \in \mathbb{R}^{N_t^2 \times S_t}
\]
and
\[
\mathbf{A}^i = [\text{vec}(\mathbf{A}_{t_i+1}), ..., \text{vec}(\mathbf{A}_{t_i+s})] \in \mathbb{R}^{N_t^2 \times S_t}
\]
and all these outputs will be stored into MATLAB cells of size $N\times 1$ as 
\[
\mathcal{C} = \{\mathbf{C}^1, ..., \mathbf{C}^N\}
\]
and
\[
\mathcal{A} = \{\mathbf{A}^1, ..., \mathbf{A}^N\}.
\]

\subsection{Network Architecture}
\subsubsection{Architecture 1:} This is a individual sequence training model. In this setting, we will construct a LSTM architecture accepting the sequences of $\mathbf X^i$ as input for training purpose. The training process of each sequence is independent. Meanwhile, during training, a single column of $\mathbf X^i$ will be sent into the network architecture at each time step. Hence, in each $t$, the data dimension is $D_t N_t \times 1$. As each target in the target sequence is in dimension $N_t N_t \times 1$, directly taking the output from the LSTM to the target will need a hidden dimension of $N_t N_t \times 1$. This will produce huge numbers of weight parameters in LSTM at scale of $O(D_t\times N_t^3)$. We will use an LSTM of a much smaller hidden input size $h$ and then use a fully connected layer to connect LSTM with the target.

As constrained in \eqref{Eq:Ac2}, we do not expect the network to produce the diagonal elements of each $\mathbf C_t \in\mathbb{R}^{N_t\times N_t}$. Therefore, the number of entries of the output from the fully connected layer should be $N_t^2-N_t$.  To match MATLAB requirement on the Regression Layer, we will custom a padding layer to add zeros back to the diagonal of $\mathbf{C}_t$ to make sure $\text{diag}(\mathbf C_t) = 0$. To this end, we have defined two custom layers: one for padding and one for loss. Plus, the new custom regression layer will take a parameter for $\lambda > 0$.

The algorithm can be summarized as  {Algorithm \ref{Alg:1}}.

\subsubsection{Architecture 2:} This is a sequence-to-sequence training model, and we will call it "Solver" in the following experiment section. In this training, the working mechanism is mostly the same with Architecture 1, except that the model accept all sequences of $\mathcal{X}_{\text{train}}$ and $\mathcal{Y}_{\text{train}}$ as input and output separately and train them with one model in a whole.

The algorithm can be summarized as {Algorithm \ref{Alg:2}}.

The networks for both architectures can be defined in MATLAB code as

\texttt{layers = [ ...}

\texttt{    sequenceInputLayer($D_t*N_t$) }

\texttt{   lstmLayer($h$, 'OutputMode', 'sequence') }

\texttt{   fullyConnectedLayer($N_t*N_t - N_t$) }

\texttt{   myPaddingLayer($N_t$) }

\texttt{   myRegressionLayer('Evolving', $\lambda$)]; }

\begin{algorithm} 
\renewcommand{\algorithmicrequire}{\textbf{Input:}}
\caption{Architecture 1 of LSTM-ESCM Algorithm}
\label{Alg:1}
\begin{algorithmic}[1]
\REQUIRE Training dataset $\mathcal{X}_{\text{train}}$. Target dataset $\mathcal{Y}_{\text{train}}$. Regulariser $\lambda$. Hidden size $h$. Learning rate $\eta$.
\FOR{$i=1, 2, ..., N$}
\STATE Apply (\ref{Eq:Ac2}), (\ref{Eq:B2}), and  (\ref{Eq:B3}) to conduct clustering training with LSTM networks and generate $\mathbf{C}^i$;

\STATE Construct affinity matrix $\mathbf{A}^i$ from $\mathbf{C}^i$; 

\STATE Implement spectral clustering to $\mathbf{A}^i$ to obtain data segmentation;
\ENDFOR
\end{algorithmic}
\end{algorithm}

\begin{algorithm} 
\renewcommand{\algorithmicrequire}{\textbf{Input:}}
\caption{Architecture 2 of LSTM-ESCM Algorithm}
\label{Alg:2}
\begin{algorithmic}[1]
\REQUIRE Training dataset $\mathcal{X}_{\text{train}}$. Target dataset $\mathcal{Y}_{\text{train}}$. Regulariser $\lambda$. Hidden size $h$. Learning rate $\eta$.

\STATE Apply (\ref{Eq:Ac2}), (\ref{Eq:B2}), and  (\ref{Eq:B3}) to conduct clustering training with LSTM networks on all sequences 
and generate $\mathcal{C}$;

\FOR{$i=1, 2, ..., N$}
\STATE Construct affinity matrix $\mathbf{A}^i$ from $\mathbf{C}^i$; 

\STATE Implement spectral clustering to $\mathbf{A}^i$ to obtain data segmentation;
\ENDFOR

\end{algorithmic}
\end{algorithm}

\section{The Experimental Results} \label{Sec:5}
In this Section, we investigate the performance of our proposed model. 
All experiments are carried out on a laptop machine running a 64-bit operating system with Intel Core i7-7660U CPU @ 2.50GHz and 16G RAM with MATLAB 2019a version. 

\subsection{Real Time Motion Segmentation Tasks}
The motion segmentation problem - recovering scene geometry and camera motion from a sequence of images - is a very crucial pre-processing step for multiple applications in computer vision, such as surveillance, tracking, action recognition, etc, and has attracted much of the attention of the vision community over the last decade \cite{tron2007benchmark,tomasi1992shape}. The problem is formed as clustering a set of two dimensional trajectories extracted from a video sequence with several rigidly moving objects into groups; the resulting clusters represent different spatial-temporal regions. 

The video sequence is often times obtained as a stream of frames and it is mostly processed in a real-time mode \cite{smith1995asset}. In the real-time setting, the $t$th snapshot of $\mathbf{X}_t$ (a time interval consisting of multiple video frames) is of
dimension $2F_t\times N_t$, where $N_t$ is the number of trajectories at $t$th time interval, $F_t$ is the number of video frames received in $t$th time interval, $n_t$ is the number of rigid motions at $t$th time interval, and $F=\sum_t F_t$ denotes the total number of frames \cite{hashemi2018evolutionary}. 

As the obtained video sequence is identified by its temporal pattern, the real-time motion segmentation problem has a good chance of getting solved by evolutionary subspace clustering algorithms. Specifically, the trajectories of $n_t$ rigid motions sit in a set of $n_t$ low-dimensional subspaces in $\mathbb{R}^{2F_t}$ at $t$th snapshot, each having no more than $d_t = 3n_t$ dimension \cite{tomasi1992shape}. In contrast, offline evolutionary subspace clustering is conducted on the whole video sequence, and thus we will expect much higher accuracy than that in the online structures. Whereas, offline modes solely limit in several specific situations and cannot be well extended to the settings where a few motions gradually vanish or new motions come up in the video sequence \cite{hashemi2018evolutionary}. To validate the performance of the proposed LSTM-ESCM framework, Hopkins 155 database \cite{tron2007benchmark} is considered here. 

\subsection{Hopkins 155 Dataset} \label{Sec:5_1}
The database collects $50$ video sequences of indoor and outdoors scenes containing two or three motions, in which each video sequence $\mathbf{X}$ with three motions was split into three motion sequences $\mathbf{X}_\text{g12}$, $\mathbf{X}_\text{g13}$ and $\mathbf{X}_\text{g23}$ containing the points from groups one and two, one and three, and two and three, respectively. This gives a total of 155 motion sequences: $120$ with two motions and $35$ with three motions, in which the number of checkerboard sequences is $104$, traffic is $38$, and articulated/non-rigid is $13$. 

Checkerboard sequences are indoor scenes taken with a handheld camera under controlled conditions. The checkerboard pattern on the objects is used to assure a large number of tracked points. Traffic scenes are taken by a moving handheld camera and most of them contain degenerate motions, particularly linear and planar motions. Articulated/non-rigid sequences display motions constrained by joints, head and face motions, people walking, etc \cite{tron2007benchmark}.

\begin{figure*}[tbh]
\centering
\subfigure[1R2RC]{ \label{fig:1R2RC}
\includegraphics[height=0.15\linewidth,width=0.20\linewidth]{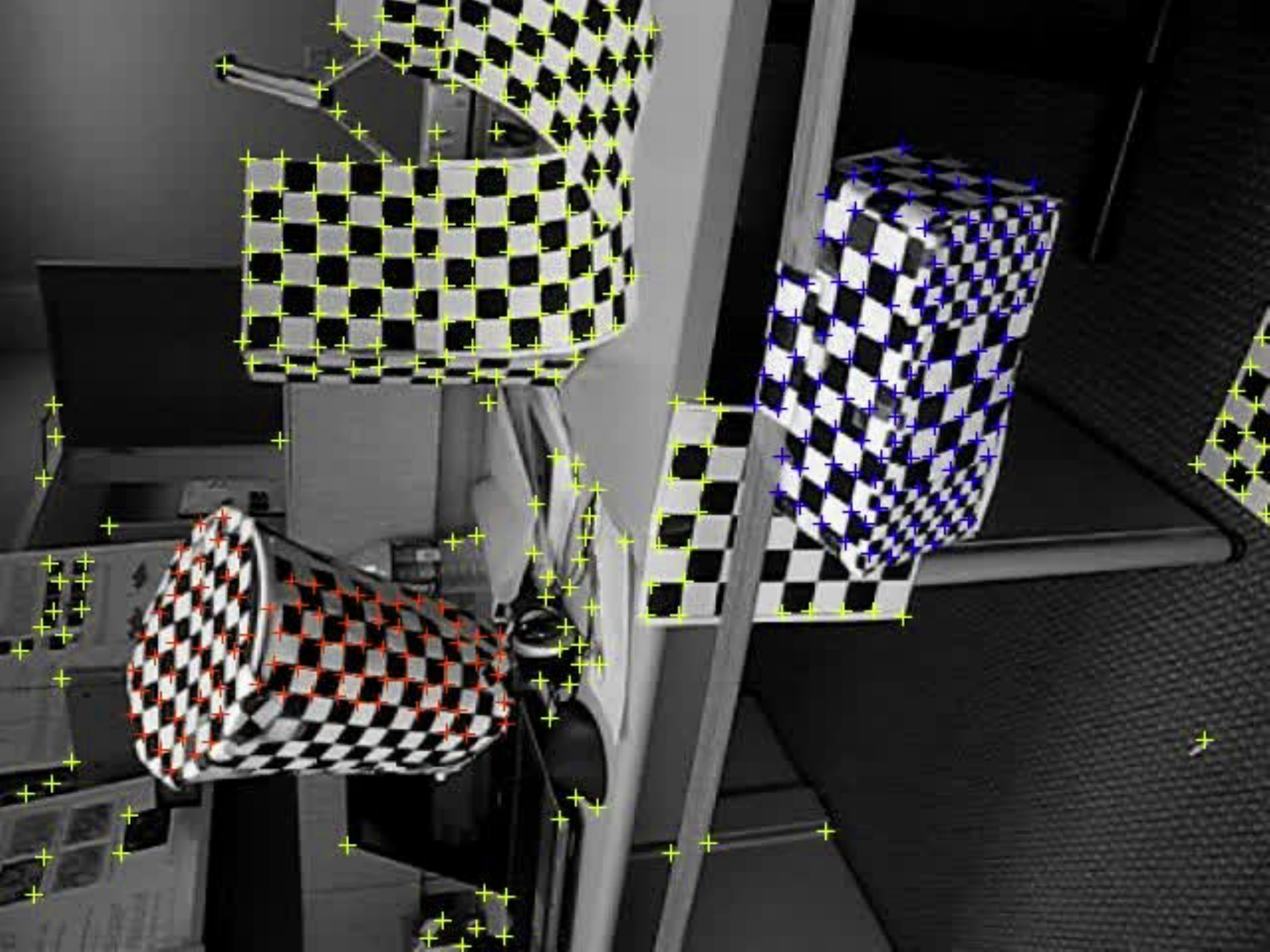} }
\subfigure[1R2RCR]{\label{fig:1R2RCR}
\includegraphics[height=0.15\linewidth,width=0.20\linewidth]{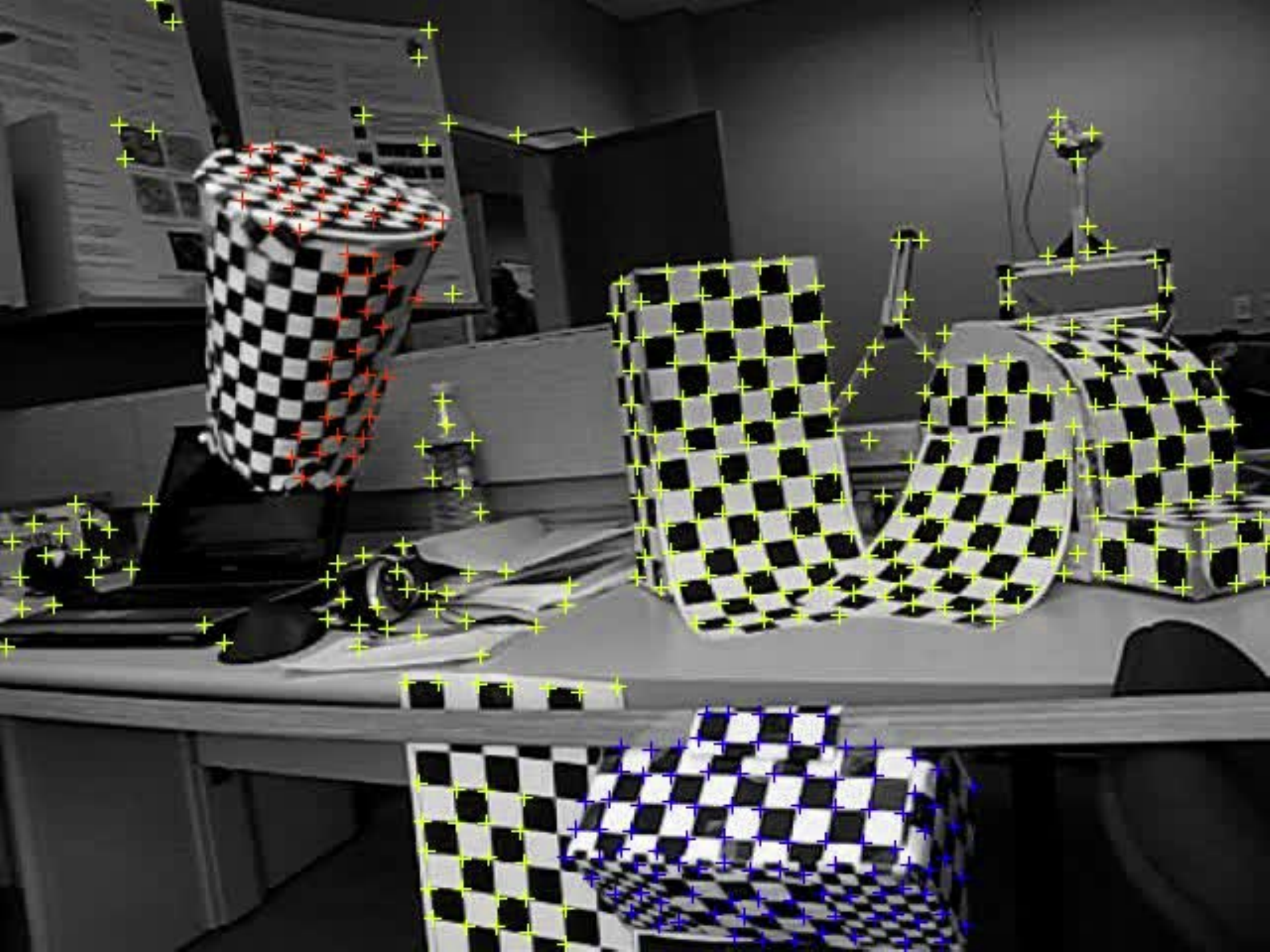}}
\subfigure[1R2RCTA]{\label{fig:1R2RCT_A} 
\includegraphics[height=0.15\linewidth,width=0.20\linewidth]{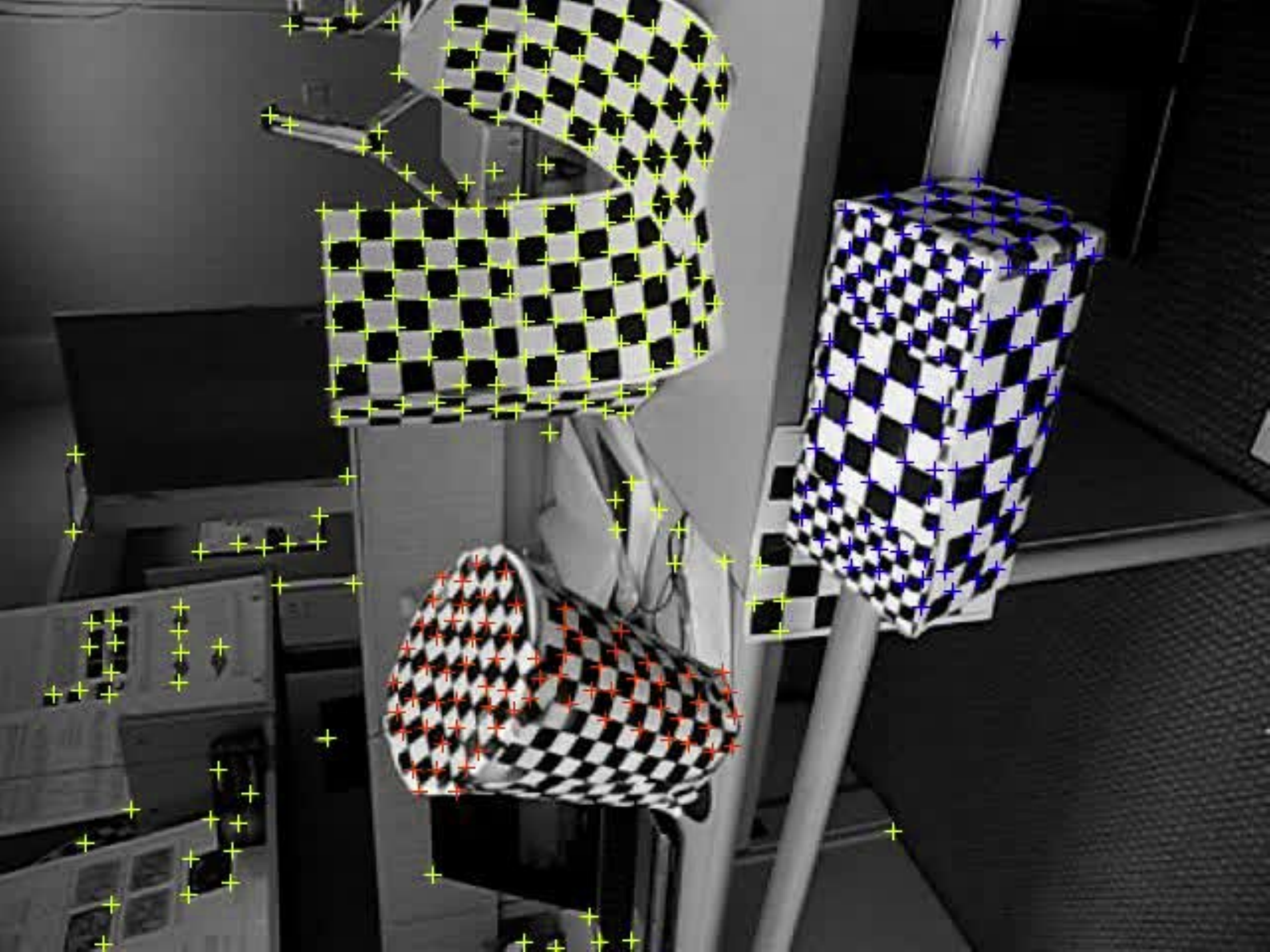}}
\subfigure[1R2RCTB]{\label{fig:1R2RCT_B} 
\includegraphics[height=0.15\linewidth,width=0.20\linewidth]{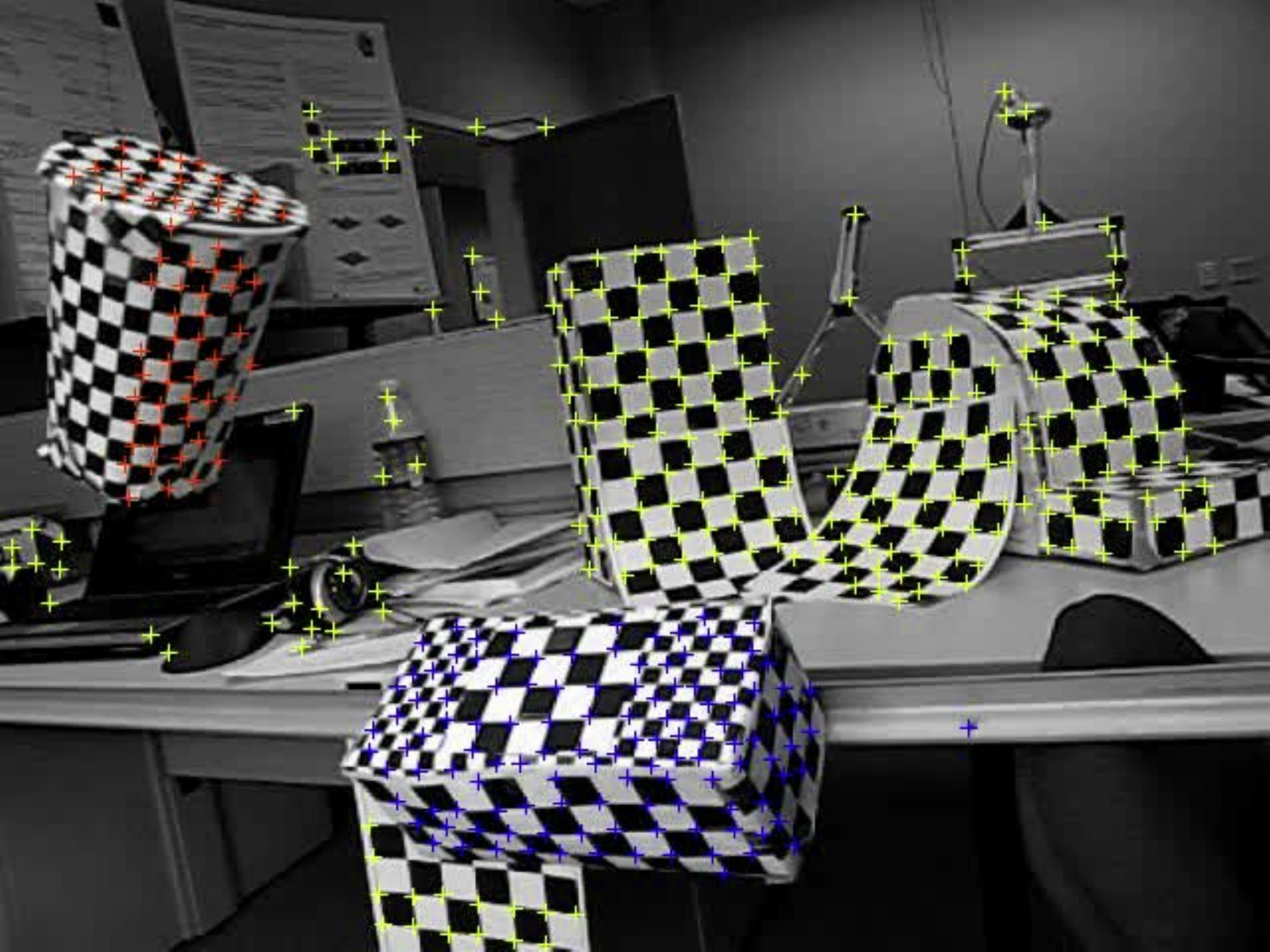}}

\subfigure[cars1]{ \label{fig:cars1}
\includegraphics[height=0.15\linewidth,width=0.20\linewidth]{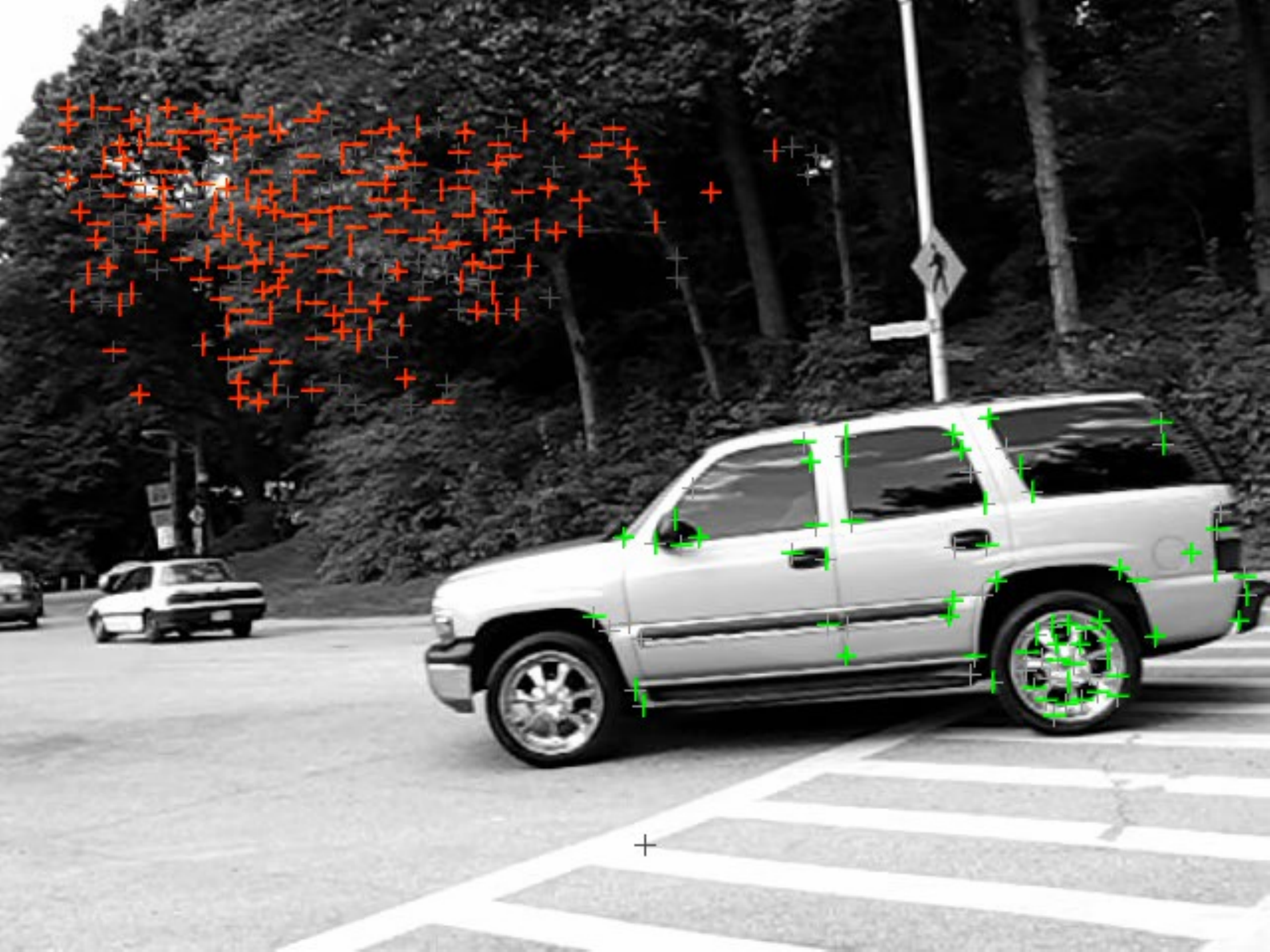} }
\subfigure[car2]{\label{fig:car2}
\includegraphics[height=0.15\linewidth,width=0.20\linewidth]{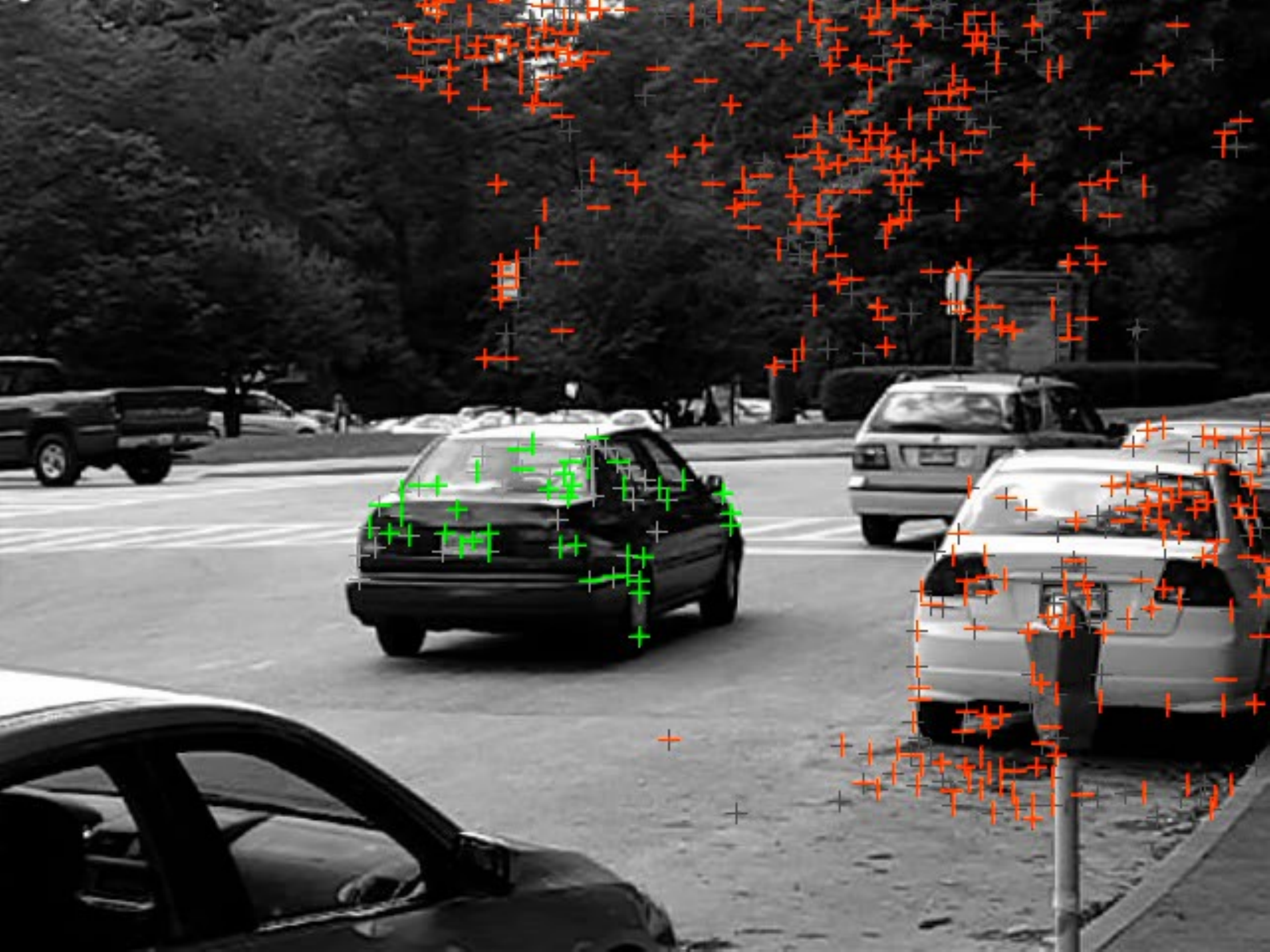}}
\subfigure[cars2.06]{\label{fig:cars2_06} 
\includegraphics[height=0.15\linewidth,width=0.20\linewidth]{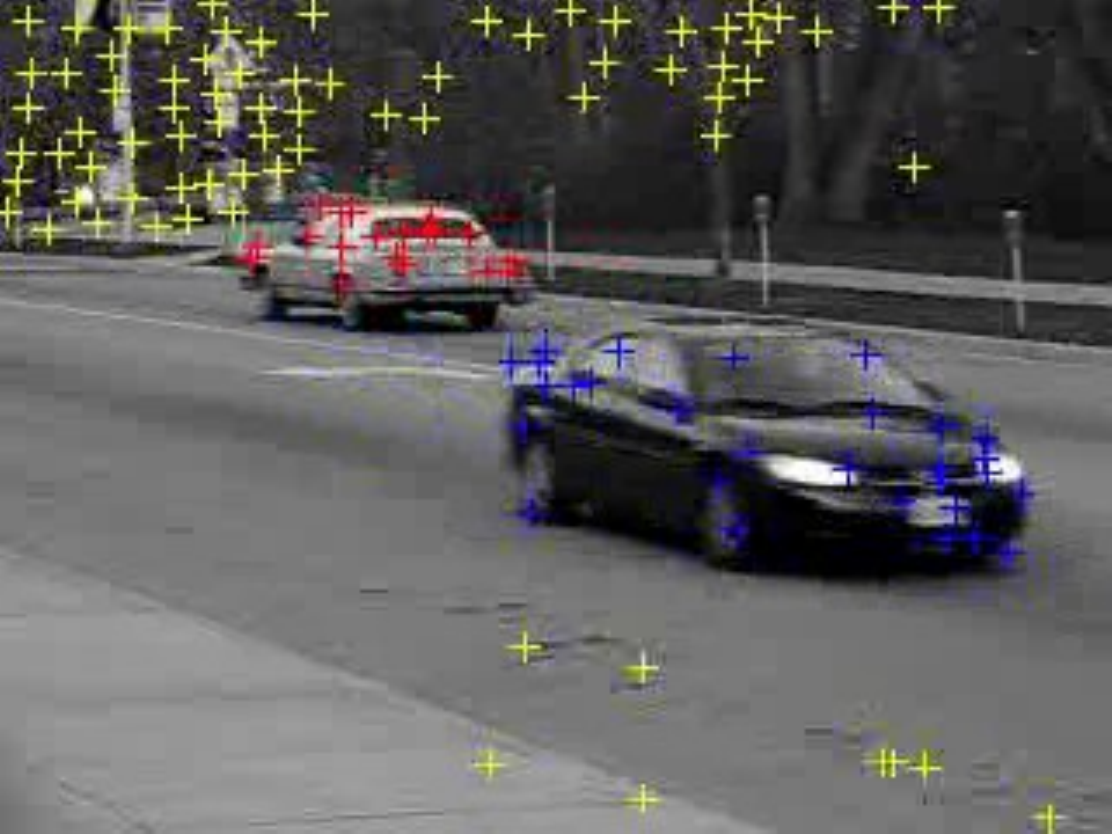}}
\subfigure[cars10]{\label{fig:cars10} 
\includegraphics[height=0.15\linewidth,width=0.20\linewidth]{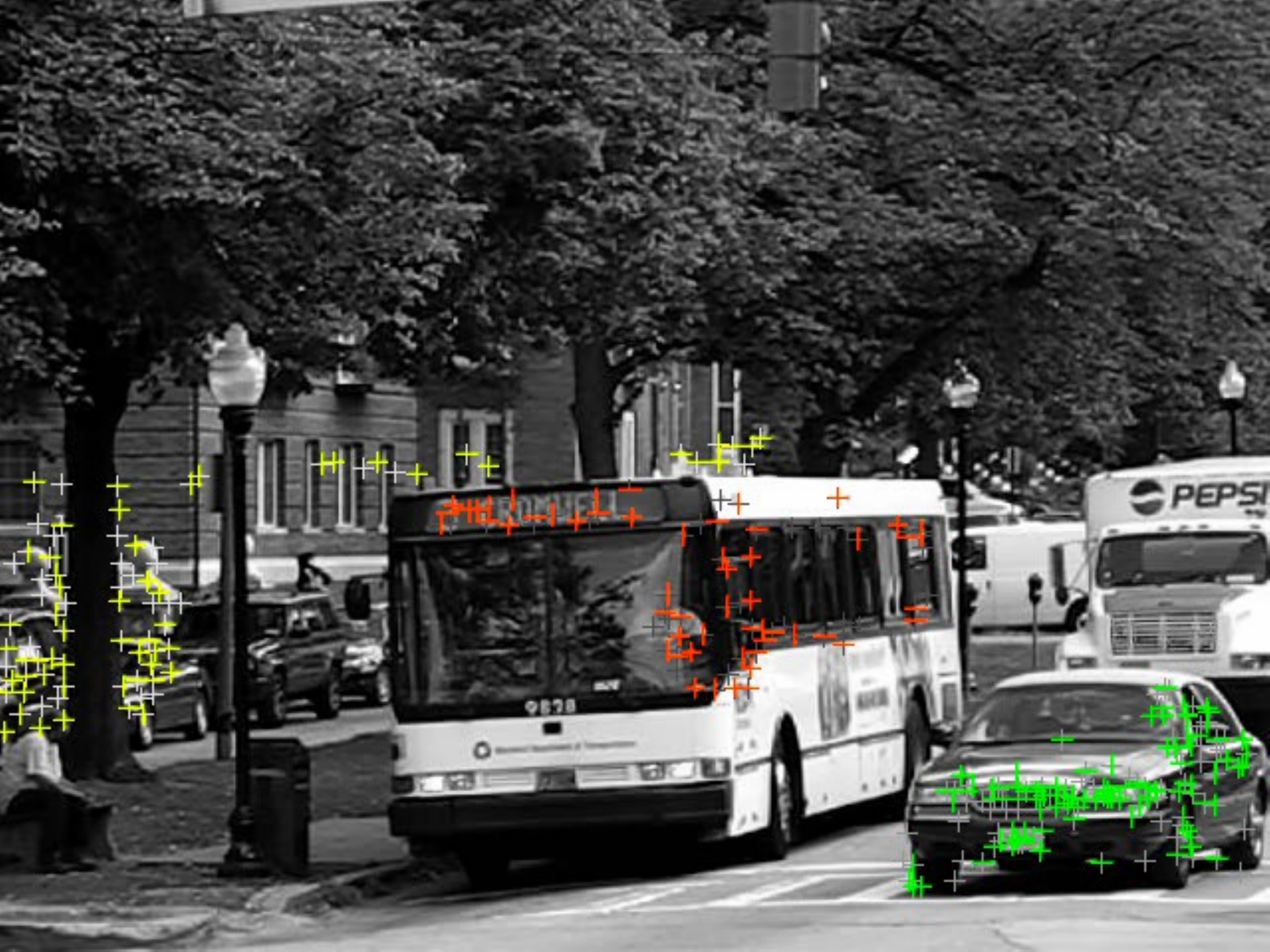}}

\subfigure[arm]{ \label{fig:arm}
\includegraphics[height=0.15\linewidth,width=0.20\linewidth]{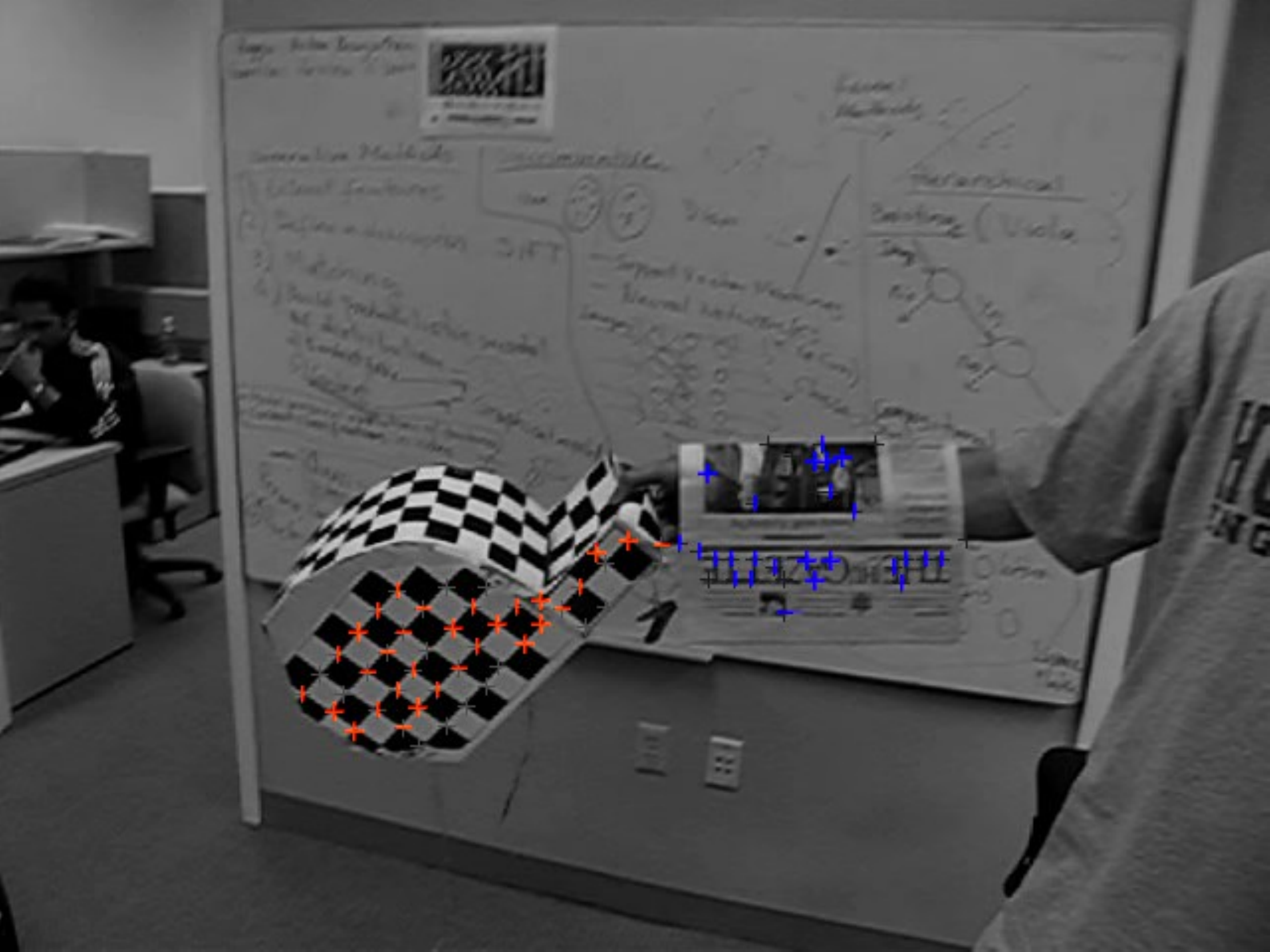} }
\subfigure[truck1]{\label{fig:truck1}
\includegraphics[height=0.15\linewidth,width=0.20\linewidth]{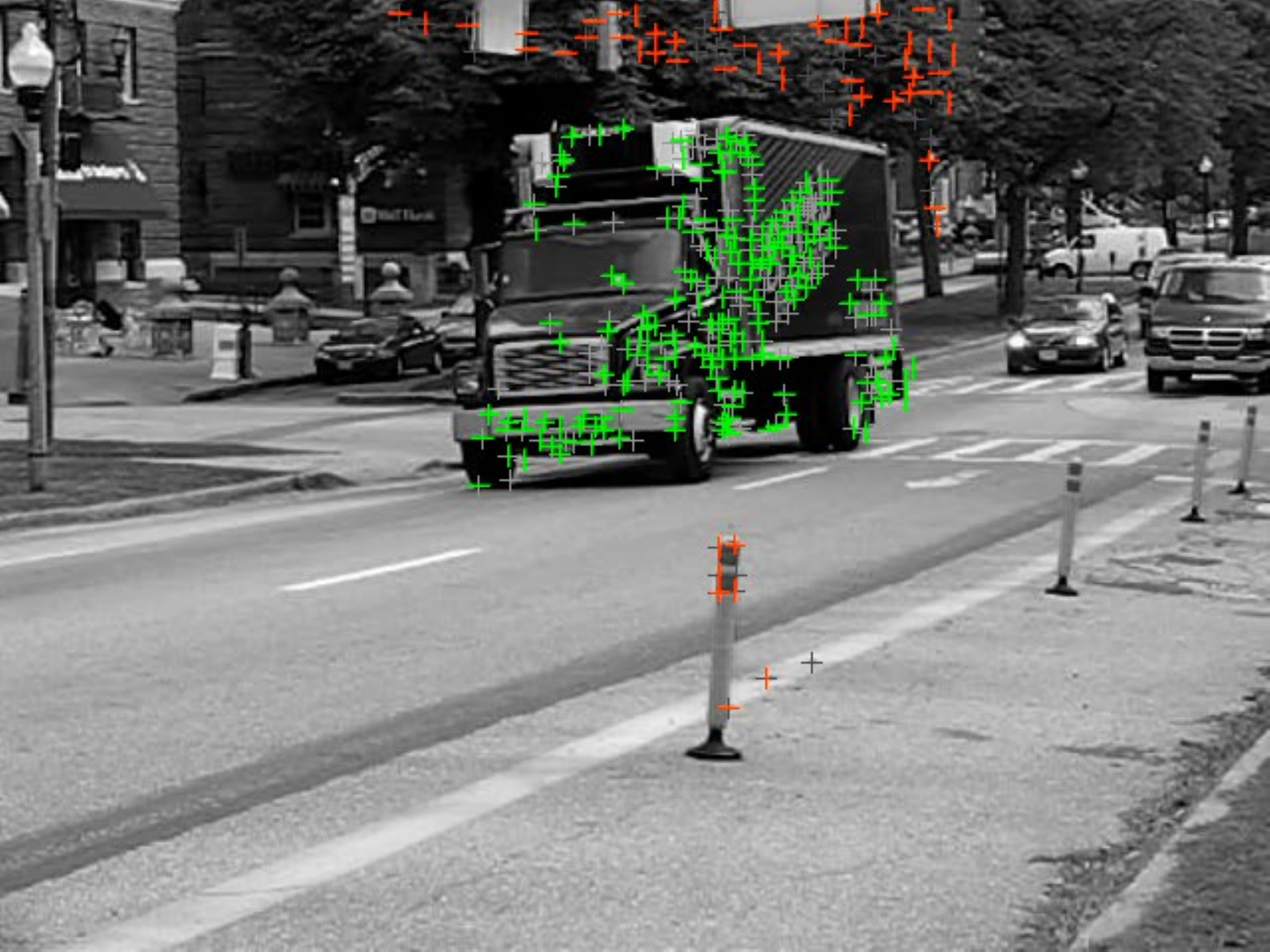}}
\subfigure[kanatani1]{\label{fig:kanatani1} 
\includegraphics[height=0.15\linewidth,width=0.20\linewidth]{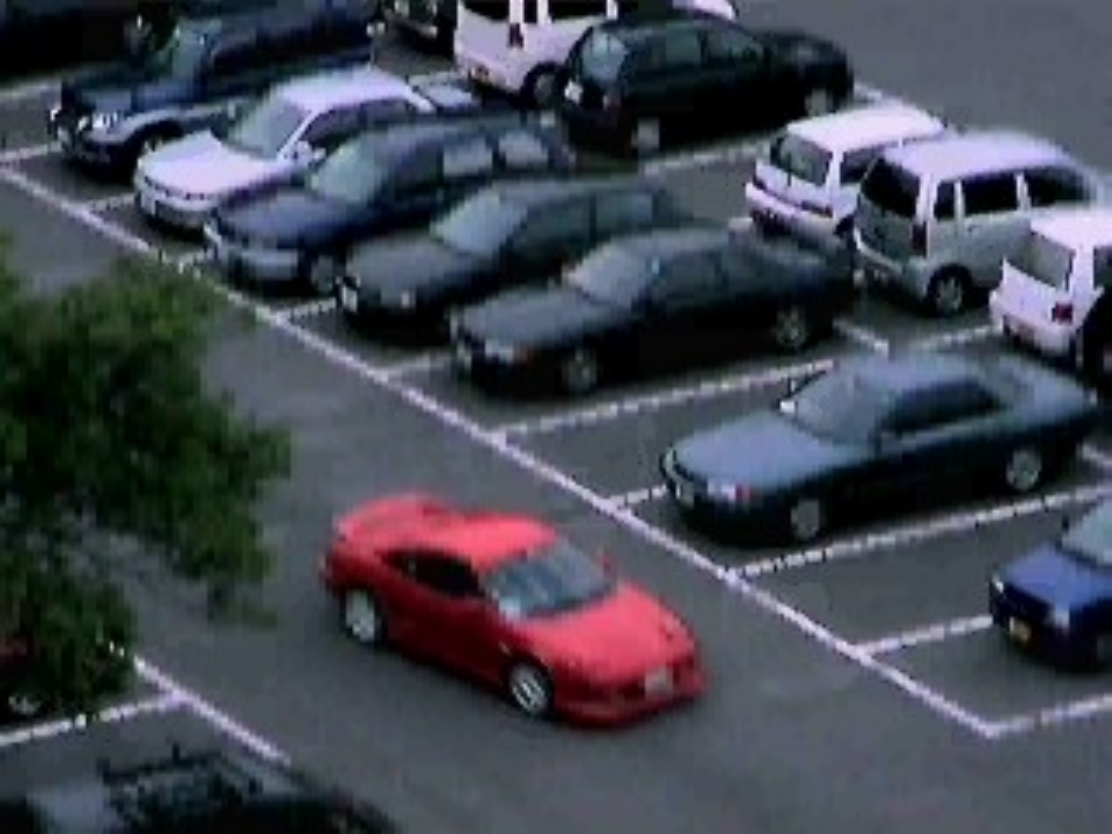}}
\subfigure[people1]{\label{fig:people1} 
\includegraphics[height=0.15\linewidth,width=0.20\linewidth]{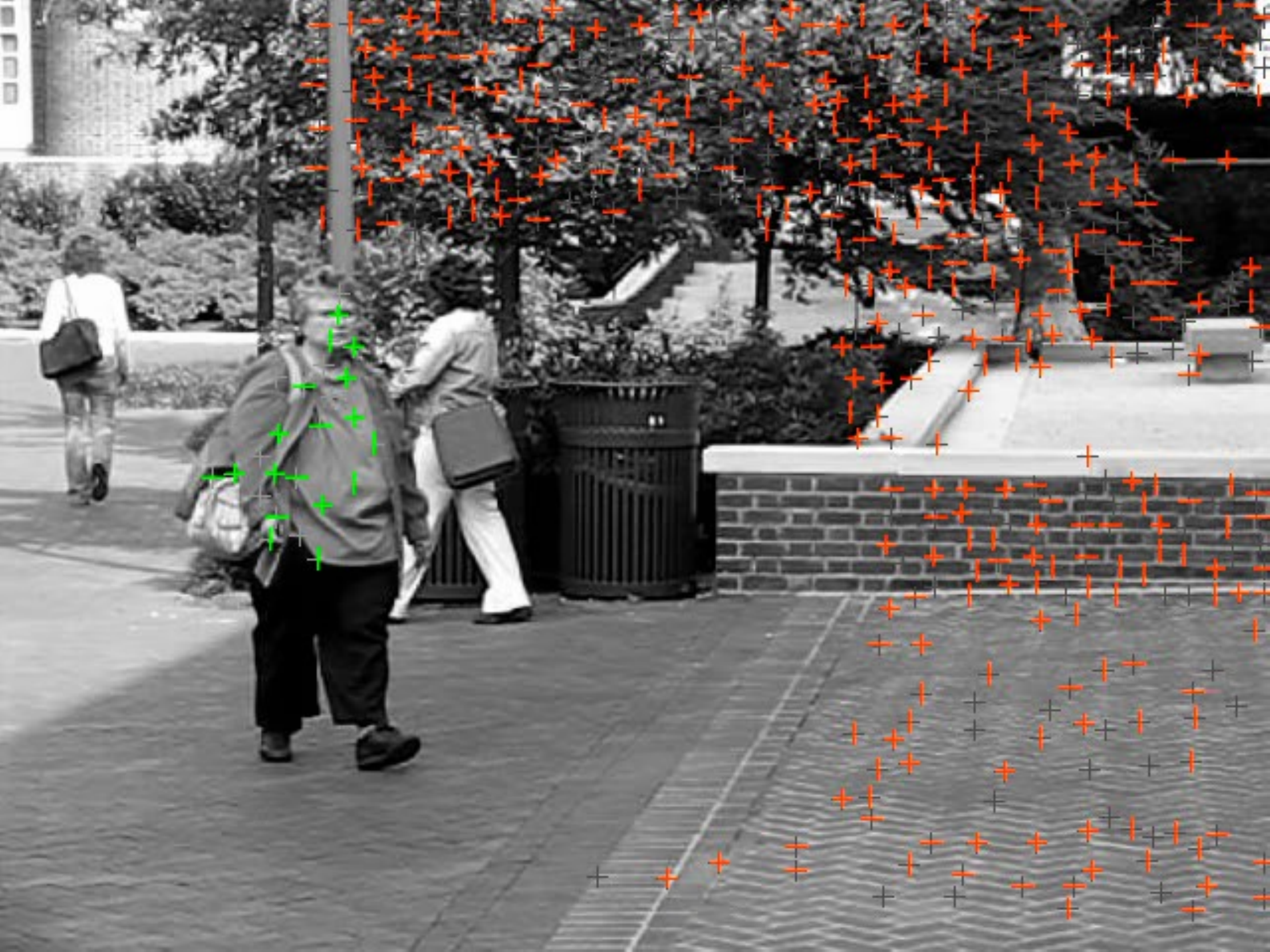}}
\caption{Sample images from some sequences in the
database with tracked points superimposed.}
\label{figure_1}
\end{figure*}

The entire database is available at \cite{jhuCVML}. {Table \ref{table_1}} reports the number of sequences and the average number of tracked points and frames for each category. In the entire dataset, per sequence, the number of points ranges from 39 to 556, while frames from 15 to 100. Point distribution represents the average distribution of points per moving object, where the last group corresponds to the camera motion (motion of the background). The statistic in table is solely computed based on the original 50 videos \cite{tron2007benchmark}.  Example frames from the videos in the Hopkins 155 dataset are shown in Figure~\ref{figure_1}.

\begin{table}
\centering
\caption{Distribution of the number of points and frames.}
\label{table_1}
\begin{tabular}{c|ccc|ccc} 
\hline
             & \multicolumn{3}{c|}{2 Groups} & \multicolumn{3}{c}{3 Groups}  \\
             & \# Seq. & Points & Frames     & \# Seq. & Points & Frames     \\ 
\hline
Check.       & 78      & 291    & 28         & 26      & 437    & 28         \\
Traffic      & 31      & 241    & 30         & 7       & 332    & 31         \\
Articul.     & 11      & 155    & 40         & 2       & 122    & 31         \\
All          & 120     & 266    & 30         & 35      & 398    & 29         \\ 
\hline
Point Distr. & \multicolumn{3}{c|}{35-65}    & \multicolumn{3}{c}{20-24-56}  \\
\hline
\end{tabular}
\end{table}

\subsection{Experiment 1: Individual Sequence Motion Segmentation} \label{Seq:5_2}

\begin{table*} 
\centering
\caption{Performance comparisons of static subspace clustering, AFFECT and CESM benchmark algorithms on Hopkins 155 dataset}
\label{table_2}
\begin{tabular}{c|cc|cc|cc} 
\toprule
\multirow{2}{*}{Learning Method} & \multicolumn{2}{c|}{Static} & \multicolumn{2}{c|}{AFFECT} & \multicolumn{2}{c}{CESM}  \\
                                 & error (\%) & runtime (s)      & error (\%) & runtime (s)      & error (\%) & runtime (s)    \\ 
\hline
BP                               & 10.76    & 1.92             & 9.86     & 1.29             & 8.77     & 1.32           \\
OMP                              & 31.66    & 0.06             & 14.47    & 0.03             & 6.85     & 0.03           \\
AOLS ($L=1$)                     & 16.27    & 0.27             & 9.27     & 0.20             & 8.24     & 0.22           \\
AOLS ($L=2$)                     & 8.54     & 0.5              & 6.17     & 0.23             & 5.70     & 0.25           \\
AOLS ($L=3$)                     & 6.97     & 0.76             & 5.92     & 0.26             & 5.60     & 0.28           \\
\toprule
\end{tabular}
\end{table*}

\begin{table*} 
\centering
\caption{Performance of proposed framework on Hopkins 155 dataset}
\label{table_3}
\begin{tabular}{c|cc|c|c} 
\toprule
\multirow{2}{*}{Learning Method} & \multicolumn{2}{c|}{smoothing} & Test on 1 snapshot & Test on 2 snapshots  \\
                                 & error (\%) & runtime (s)         & error (\%)           & error (\%)             \\ 
\hline
Proposed LSTM-ESCM                    & 0.005    & 0.13                & 4.42               & 4.42                 \\
\toprule
\end{tabular}
\end{table*}

In contrast to most of the previous work which handles the aforedescribed dataset in an offline mode, we will process it in a real-time setting. Each video is divided into $T$ data matrices $\{\mathbf{X}_t\}_{t=1}^T$ so that $F_t \ge 2n$ for a video sequence with $n$ motions. Next, PCA is applied on $\mathbf{X}_t$ and the top $D=4n$ singular vectors are kept as the final input to the representation learning algorithms. The input parameters for LSTM-ESCM are set as $\lambda = 0.1$, $h = \text{ceil}(\frac{N_t}{5})$ and $\eta = 0.001$ for each sequence. Note that $N_t$ is the same for each sequence.
 
A specific static subspace clustering algorithm \cite{vidal2011subspace}, AFFECT \cite{xu2014adaptive} ,and CESM \cite{hashemi2018evolutionary} are used as benchmarks for our proposed framework. Static subspace clustering applies subspace clustering at each time step independently from the previous outcomes; AFFECT and CESM both applies spectral clustering \cite{ng2002spectral} on the weighted average of affinity matrices $\mathbf{A}_t$ and $\mathbf{A}_{t-1}$. The default choices for the affinity matrix in AFFECT are the negative squared Euclidean distance or its exponential form, while the affinity matrices in CESM are explored under the self-expressive properties \cite{elhamifar2009sparse} of $\mathbf{X}$. Under its original settings, AFFECT achieves a clustering error of $44.1542\%$ and $21.9643\%$ using the negative squared Euclidean distance or its exponential form, respectively, which, as presented in {Table \ref{table_2}}, is inferior even to the static subspace clustering algorithms. Hence, to fairly compare the performance of different evolutionary clustering strategies, BP \cite{elhamifar2009sparse,chen2001atomic}, OMP \cite{dyer2013greedy}, and AOLS ($L = 1, 2, 3$) \cite{hashemi2017accelerated,hashemi2016sparse} are employed to learn the representations for all the benchmark schemes, including AFFECT. Plus, the results are averaged over all sequences and time intervals excluding the initial time interval $t = 1$. The initial time interval is excluded because for a specific representation learning method (e.g., BP), the results of static as well as evolutionary subspace clustering coincide \cite{hashemi2018evolutionary}.

The performances of multiple benchmark schemes are presented in {Table \ref{table_2}}, where all the error rates are smoothing (training) errors, while the performances of our proposed LSTM-ESCM framework are shown in {Table \ref{table_3}}, in which both smoothing errors as well as test errors on the last one or two snapshots of each sequence are displayed. Note that for the CESM algorithm, though authors in \cite{hashemi2018evolutionary} list results corresponding to both a constant smoothing factor and a smoothing factor achieved by using their proposed alternating minimization schemes, we will solely keep the later one as our benchmark.

As presented in {Table \ref{table_2}}, for all the representation learning schemes, static subspace clustering has the highest clustering errors among all benchmark models due to not incorporating any knowledge about the representations of the data points at other times, while CESM performs relatively better than its AFFECT counterparts. For runtimes, methods rooted on OMP are the most time-efficient, whereas BP-based approaches have the slowest processing speed. The outcomes in {Table \ref{table_3}} show that the cluster errors of LSTM-ESCM is fairly promising and the training error is not even in the same scale as those in {Table \ref{table_2}} and meanwhile even the test clustering outcomes perform better than the training error rates of all our benchmark models. Plus, apart from a bit slower than methods based on the OMP optimization regime, the runtime of our proposed framework is also quite superior. 

\subsection{Experiment 2: Overall Motion Segmentation Solver}
In the previous experiment, we explored the performance of the LSTM-ESCM model in segmenting motion sequence. Each sequence has its own evolving LSTM-ESCM model. 

In the following experiment, we want to explore whether we can build up a well performed LSTM-ESCM for a number of motion sequence so that the learned LSTM can be used as a universal solver for the affinity matrix $\mathbf C_t$ for any sequences from the similar video contexts. 
Our method is to train the LSTM-ESCM in a whole by using the sequence data from multiple different motion sequences. To make the training task easier, we make an assumption that the number of keypoints to be segmented is the same for all the sequence.


This experimental study is still set on the Hopkins Motion 155 dataset. Since, in this set, the number of keypoints for each sequence varies from $63$ to $548$, whereas our assumption for building the overall LSTM-ESCM solver is under the setting of uniform number of keypoints. Thus, in our attempts, $300$ keypoints are set as the target for qualified sequences. To achieve that, we first filter out $21$ specific sequences by thresholding between $300$ and $350$ keypoints in their original. Then, by limiting $300$ keypoints for all the $21$ selected ones, we randomly remove equal number of excess keypoints in all motions for each sequence. Next, $17$ sequences are randomly selected as the training set and the rest $4$ is left as test set. Finally, to this end, the LSTM-ESCM solver is built by learning the whole training set altogether. The error rate we get on test set is $32.86\%$, and the training time is $850.34s$.

\subsection{Experiment 3: Ocean Water Mass Clustering}
The Argo Program has succeeded in achieving more than 13 years of global coverage from 2004 to the present. The website 
\url{http://sio-argo.ucsd.edu/RG_Climatology.html} hosts the Argo dataset with related ocean observations from other programs. In this experiment, we will apply our evolutionary subspace clustering algorithm to analyse the ocean temperature and salinity profiles for ocean water mass clustering.

Study \cite{LiXuZhouWangWrightLiuLin2017} shows that a body of water with a common features such as salinity and temperature can be used to characterize a watermass. Analysing water masses can facilitate our understanding of global climate change, seasonal climatological variations, ocean  biogeochemistry,  and  ocean  circulation  and  its effect  on  transport  of  oxygen  and  organisms.  

To demonstrate the ability and performance of our LSTM evolutionary subspace clustering in modeling various real-world problems, we conduct an experiment on clustering water masses based on the salinity and temperature profiles. The dataset considered, downloable from the above website, tracks 
the ocean temperature and salinity by Argo ocean observatory system comprising more than 3000 floats which provide 100,000 plus profiles each year.  These floats cycle between the oceansurface and 2000m depth every 10 days, taking salinity and temperature measurements at varying depths. The dataset contains normalized monthly averages (from January 2004 and December 2016) of ocean salinity and temperature with 1 degree resolution worldwide. 

We will take a subset of data covering the location near the coast of South Africa where the Indian Ocean meets the South Atlantic, specifically the area is at at latitudes $25\degree$ South to $55\degree$ South and longitudes $10\degree$ West to $60\degree$ East.

At each gridded location, we construct a feature vector of dimension 48 consisting of (normalized) salinity and temperature values of two years (24 months) at the depth level measured at 1000 dbar. There are in total of 1684 valid gridded location. Hence at each time step $t$, the data $\mathbf X_t$ is a matrix of size $48\times 1684$. Our evolutionary sequence consists of 6 time steps, i.e., $t=1$ for Jan 2004 - Dec 2005; $t=2$ for Jan 2006 - Dec 2007; $t=3$ for Jan 2008 - Dec 2009; $t=4$ for Jan 2010 - Dec 2011;  $t=5$ for Jan 2012 - Dec 2013; and $t=1$ for Jan 2014 - Dec 2015.

\begin{figure*}[tbh]
\begin{center}
\subfigure[LSTM-ESCM]{\includegraphics[width = 0.47\textwidth]{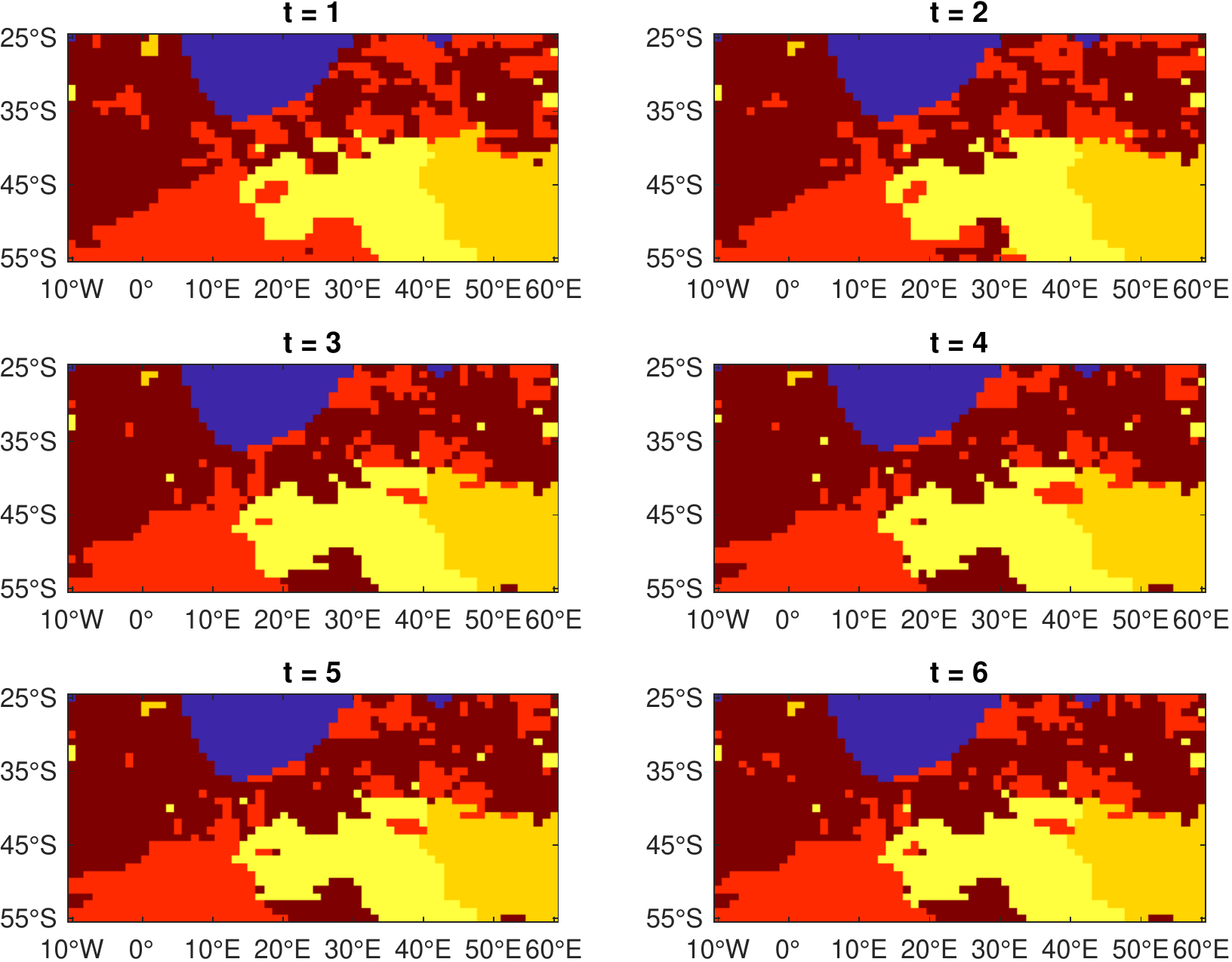}}\;\;\;
\subfigure[CESM-AOLS-L3]{\includegraphics[width = 0.47\textwidth]{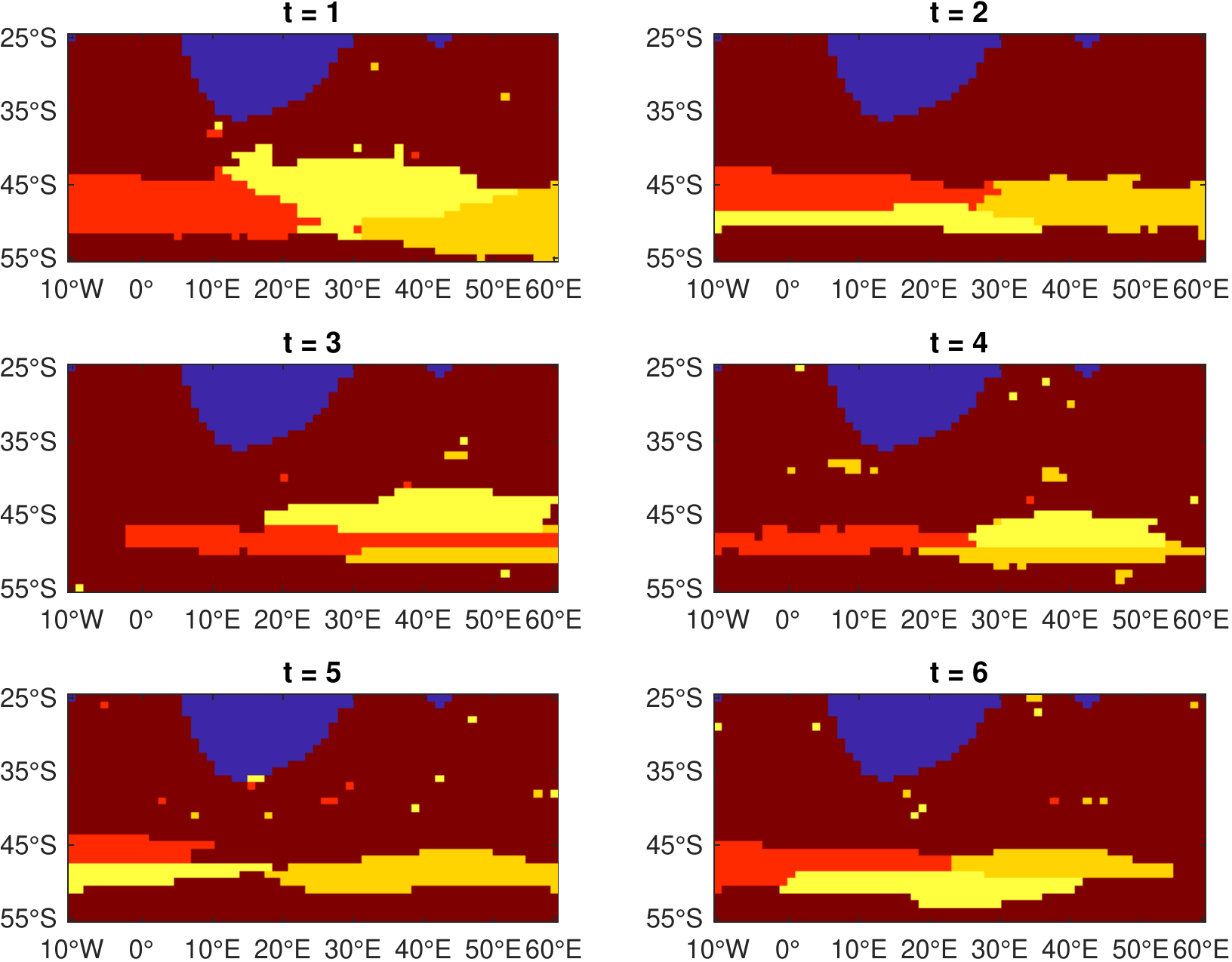}}
\end{center}
\caption{The Comparing Results from LSTM-ESCM and CESM Methods: Four types of water masses are identified by different colors. All the clustering results at six time steps from two methods are shown where $t=i$ in the figure titles show the time steps. The dark red color means the type of Other Water Masses.}\label{Fig:3}
\end{figure*}

The tunable parameters in model design and training are set to $\lambda = 0.01$, LSTM hidden size $h=200$, and the maximal epoch $T=300$. For the final clustering, we assume the number of clusters to be 4, accounting for the three  well-known  and  strong  water  masses:  (1) Agulhas currents, (2) the Antarctic intermediate water (AAIW), (3) the circumpolar deep water mass, and (4) other water masses in the area.  
Our result is shown in Figure~\ref{Fig:3}(a) while as a comparison the result by using the CESM framework from \cite{hashemi2018evolutionary} with a strategy of using the OLS-based representation learning of level 3 is shown in Figure~\ref{Fig:3}(b). The CESM model was actually conducted with a warm start from the initial affinity matrix given the static Sparse Subspace Clustering (SSC) model \cite{vidal2011subspace}. For the LSTM-ESCM, the initialization was randomly set by using MATLAB default LSTM initialization. From the experiment we have not seen any significant impact on the final results.

In terms of applications, every time when a clustering should be conducted as a new time step, CESM algorithm has to solve a new SSC problem, however for our LSTM-ESCM this is to simply conduct a forward step of LSTM-ESCM starting with the kept hidden state from the last time step. 

By comparing the clustering results of LSTM-ESCM and CESM methods, we can clearly see that LSTM-ESCM is capable of extracting more complicate patterns from the underlying data, and correspondingly will have higher chance of better clustering the water temperature and salinity.



\section{Conclusion} \label{Sec:6}
In this paper, we research on evolutionary subspace clustering, the problem of arranging a set of evolving data points which in actual fact lie in a union of low-dimensional evolving subspaces, and proposed the LSTM-ESCM framework to cope with the related applications. Our proposed model takes advantage of the self-expressive property behind data so as to learn out the parsimonious representation of data at each timestamp while using LSTM deep networks to conduct temporal information learning over the whole data sequence. Under the MATLAB deep learning toolbox, we combine its predefined sections and our customized loss functions to realize our proposed model. Then, the experiment is executed on real-world well-known datasets. 
The experimental outcomes demonstrate that, compared to the benchmark models, our model remarkably dominates in the case of both run time and accuracy.

Nonetheless, even after adding in a fully connected layer to lower down the computational cost of LSTM-ESCM, this model is for now still only capable of processing small to medium-scale datasets. Thus, how to expand our algorithm to solve the computational infeasibility of real-world high-dimensional data is of great interest.

For future research interests, there are four directions to be pointed out. First, it would be worthwhile to further expand the LSTM-ESCM scheme to other subspace clustering methods, such as the approaches relying on seeking low-rank representations, etc. Second, analysis of the theoretical
foundation of subspace clustering to interpret and analyze the performance of the proposed model would be meaningful. Next, apart from LSTM networks, other neural network architectures could also be exploited for conducting temporal smoothing. Lastly, how to apply LSTM-ESCM to other evolving learning tasks, such as the topic of variational continual learning \cite{swaroop2019improving}, is also of great interest to explore.

\bibliographystyle{IEEEtran}
\bibliography{FluteXuBib}


\begin{IEEEbiography}[{\includegraphics[width=1in,height=1.25in,clip,keepaspectratio]{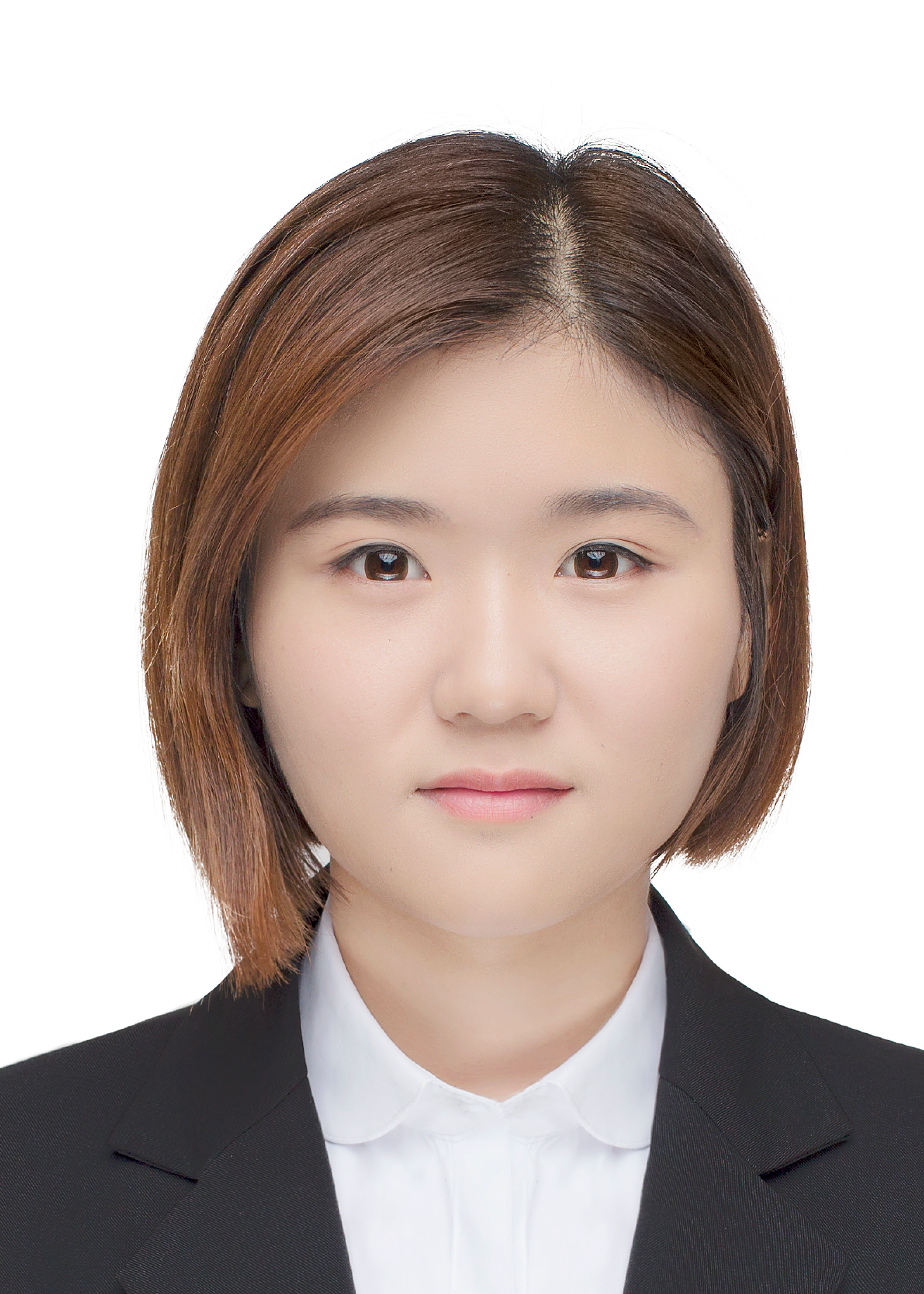}}]
{Di Xu} received MCom in business analytics degree from The University of Sydney, Australia in 2019 and B.E. in Materials in Physics from Hebei University of Technology, China in 2017.Her main research interests are in machine learning and data analytics.
\end{IEEEbiography}

\begin{IEEEbiography}[{\includegraphics[width=1in,height=1.25in,clip,keepaspectratio]{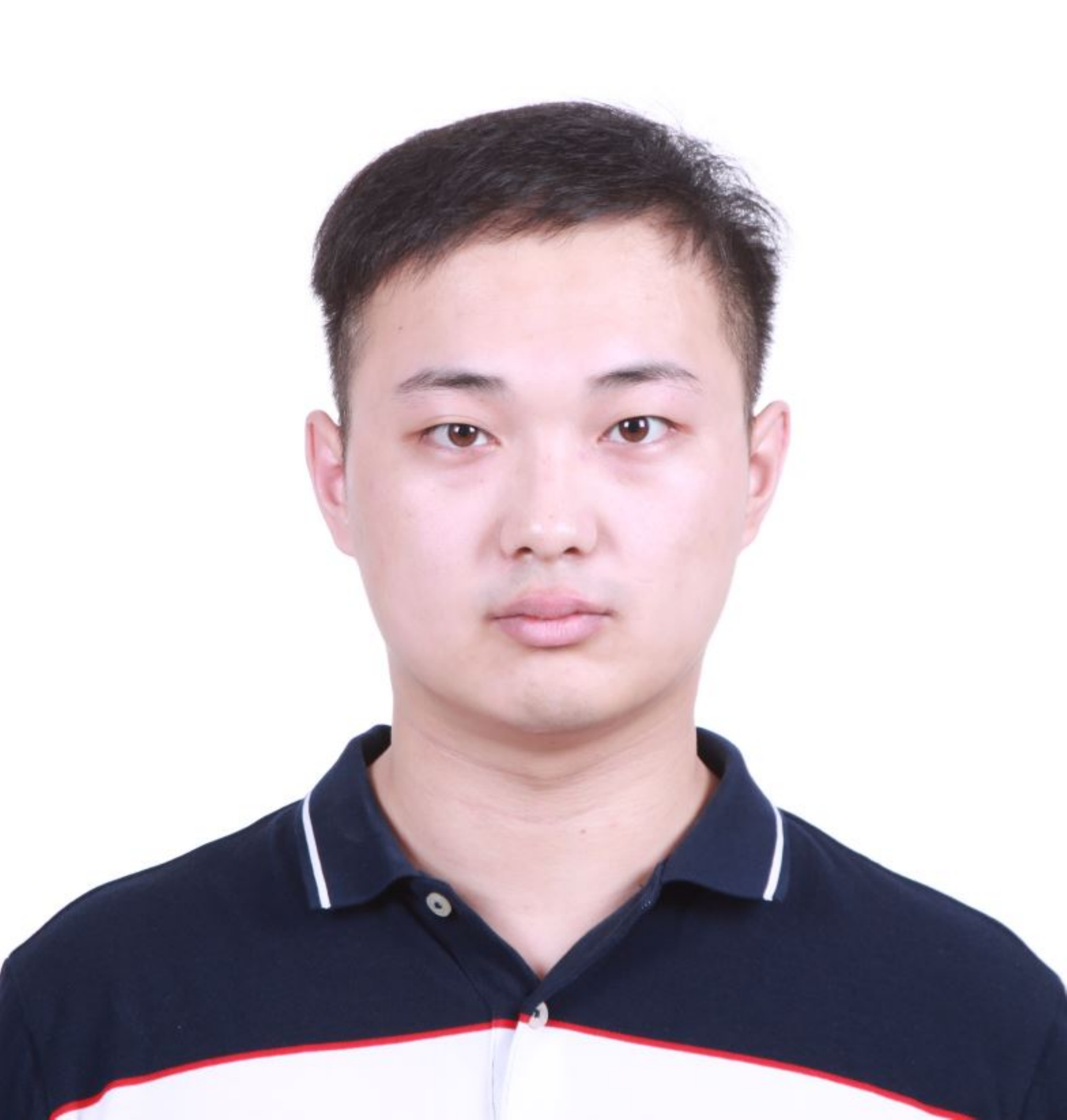}}]
{Tianhang Long} received the B.S. degrees from Beihang University, Beijing, China, in 2014. He is currently pursuing the Ph.D. degree in computer science and technology at the Faculty of Information Technology, Beijing University of Technology, Beijing, China.
His current research interests include deep learning and pattern recognition.
\end{IEEEbiography}

\begin{IEEEbiography}[{\includegraphics[width=1in,height=1.25in,clip,keepaspectratio]{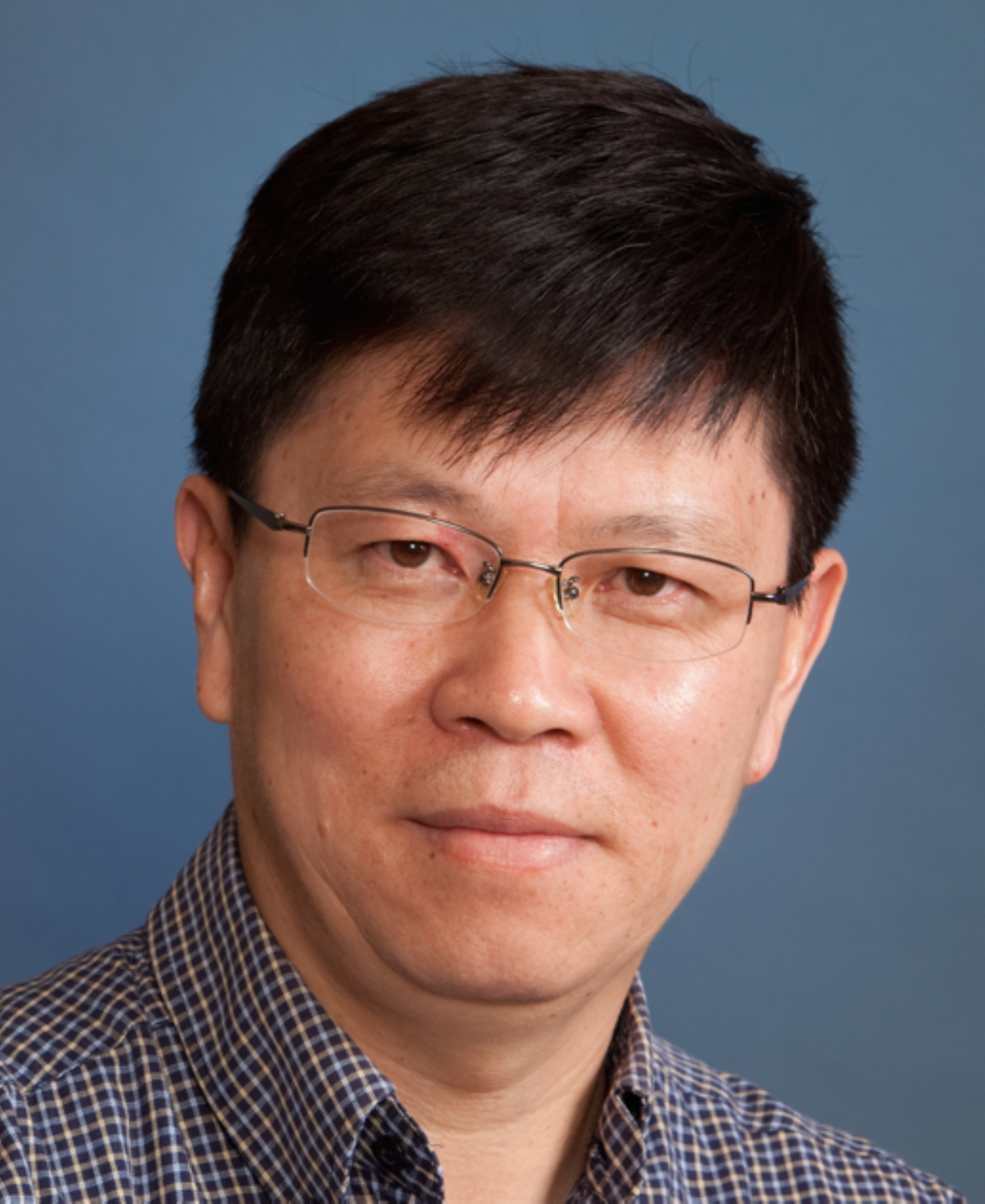}}]
{Junbin Gao} graduated from Huazhong University of Science and Technology (HUST),
China in 1982 with a BSc. in Computational Mathematics and
obtained his PhD. from Dalian University of Technology, China in 1991. He is a Professor of Big Data Analytics in the University of Sydney Business School at the University of Sydney and was a Professor in Computer Science
in the School of Computing and Mathematics at Charles Sturt
University, Australia. He was a senior lecturer, a lecturer in Computer Science from 2001 to 2005 at the
University of New England, Australia. From 1982 to 2001 he was an
associate lecturer, lecturer, associate professor, and professor in
Department of Mathematics at HUST. His main research interests
include machine learning, data analytics, Bayesian learning and
inference, and image analysis.
\end{IEEEbiography}

\end{document}